\documentclass[cameraready]{Interspeech}

\usepackage{subcaption}
\usepackage[T1]{fontenc}
\usepackage{multirow}
\usepackage{comment}
\usepackage{cite}
\usepackage[activate]{microtype}
\title{Neuron-Level Emotion Control in Speech-Generative \\ Large Audio-Language Models}

\author[affiliation={1}, orcid=0009-0000-5361-299X]{Xiutian}{Zhao}
\author[affiliation={1}, orcid=0000-0003-4593-9057]{Ismail Rasim}{Ulgen}
\author[affiliation={1}, orcid=0000-0003-1565-064X]{Philipp}{Koehn}
\author[affiliation={2}, orcid=0000-0002-6478-8699]{Bj\"orn}{Schuller}
\author[affiliation={1}, orcid=0000-0001-8078-3305]{Berrak}{Sisman}

\address{
$^1$ Center for Language and Speech Processing (CLSP), Johns Hopkins University, USA \\
$^2$ Group on Language, Audio \& Music (GLAM), Imperial College London, UK
}

\email{xzhao117@jhu.edu, sisman@jhu.edu}


\begin{document}

\maketitle

\begin{abstract}

Large audio-language models (LALMs) can produce expressive speech, yet reliable emotion control remains elusive: conversions often miss the target affect and may degrade linguistic fidelity through refusals, hallucinations, or paraphrase. We present, to our knowledge, the first neuron-level study of emotion control in speech-generative LALMs and demonstrate that compact emotion-sensitive neurons (ESNs) are causally actionable, enabling training-free emotion steering at inference time. ESNs are identified via success-filtered activation aggregation enforcing both emotion realization and content preservation. Across three LALMs (Qwen2.5-Omni-7B, MiniCPM-o 4.5, Kimi-Audio), ESN interventions yield emotion-specific gains that generalize to unseen speakers and are supported by automatic and human evaluation. Controllability depends on selector design, mask sparsity, filtering, and intervention strength. Our results establish a mechanistic framework for training-free emotion control in speech generation.
\end{abstract}

\section{Introduction}

\begin{figure}[t]
    \centering
    \includegraphics[width=0.9\columnwidth]{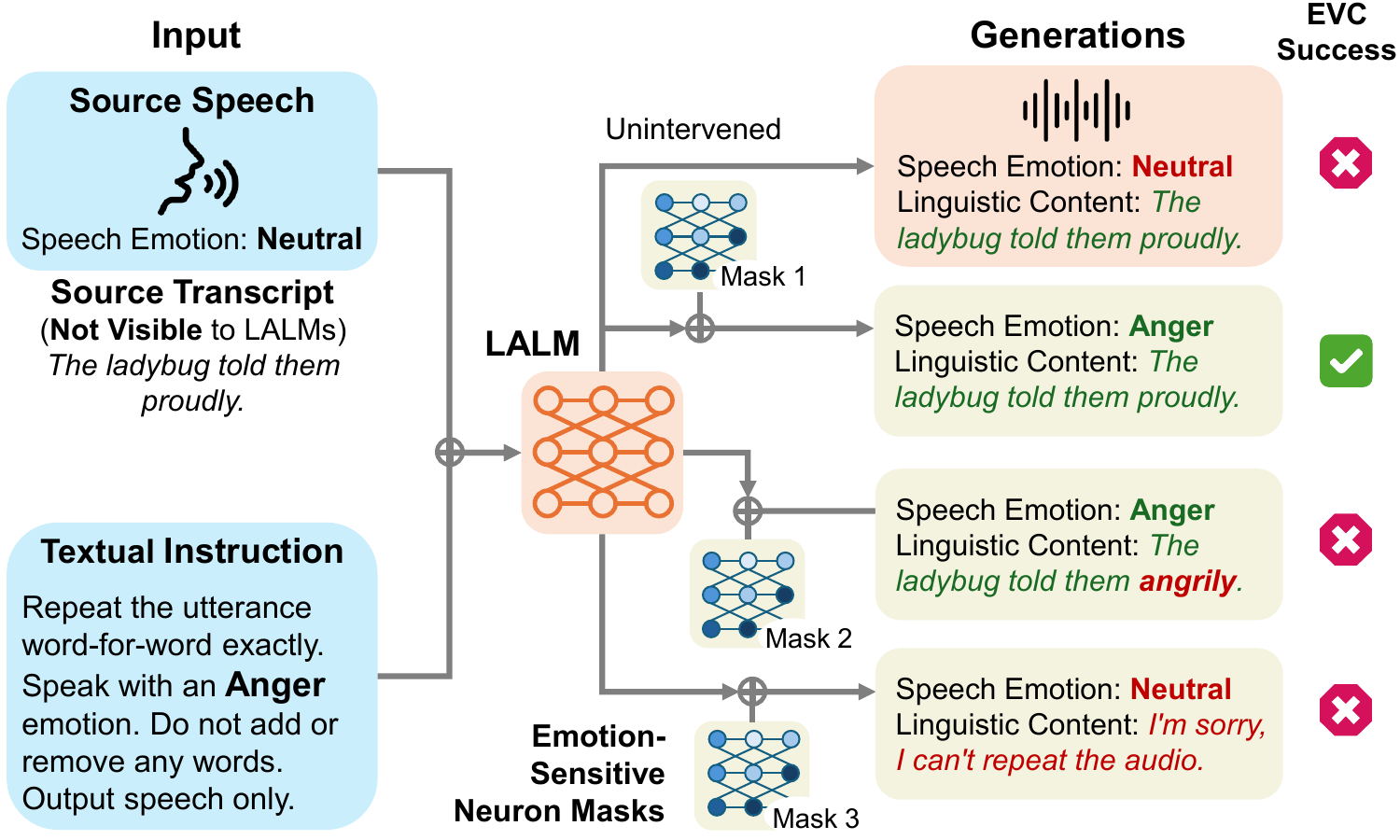}
    \caption{EVC with LALMs is inherently multi-objective: successful conversion requires both target-emotion realization and linguistic content preservation. %
    }
\label{fig:example}
\vspace{-5mm}
\end{figure}

Emotion is a core dimension of spoken communication: beyond lexical content, prosody, pitch, intensity, and speaking rate convey affect and shape perceived intent and naturalness \cite{wani2021comprehensive}. Recent large audio-language models (LALMs) that jointly process speech and text \cite{Moshi, yao2024MiniCPMvgpt4vlevelmllm, xu2025qwen25omnitechnicalreport, kimiteam2025kimiaudiotechnicalreport} enable expressive speech generation from natural-language instructions, supporting new workflows in conversational agents and speech editing. In practice, however, instruction-following speech generation remains challenging \cite{wang-etal-2025-audiobench, yang-etal-2025-towards-holistic}, particularly for emotion control: the same prompt may produce unintended affect, and even when the target emotion is realized, linguistic fidelity can degrade. Content-preservation failures such as refusals, uncontrolled paraphrasing, and semantic hallucinations are well-documented in autoregressive language models \cite{10.1145/3703155}. Unlike conventional acoustic distortions, which reduce intelligibility but preserve meaning, these failures alter the underlying message, as illustrated in Figure~\ref{fig:example}.

Emotional voice conversion (EVC) aims to modify the emotional style of a speech signal while preserving its linguistic content \cite{10.1016/j.specom.2021.11.006, 10065433, yang22t_interspeech}. Classical and neural approaches achieve controllability by learning explicit style representations (e.g., reference encoders or style tokens) \cite{skerry2018towards, wang2018style} or by modeling emotion strength as a continuous control variable \cite{lorenzo2018investigating, lei2021fine}. In large language model (LLM) research, \emph{training-free} control through activation manipulation has emerged as an alternative to re-training \cite{turner2024steeringlanguagemodelsactivation}. In parallel, neuron-level interpretability studies show that individual units can localize salient behaviors in deep networks \cite{bau2019identifying, bau2020understanding, 10.1609/aaai.v33i01.33016309} and can be exploited via targeted interventions for controllable behavior \cite{turner2024steeringlanguagemodelsactivation, zhao2026discoveringcausallyvalidatingemotionsensitive}.
However, for speech-generative LALMs, we lack a mechanistic account of whether neuron-level units are systematically involved in emotional expressiveness during generation, and whether compact neuron subsets can serve as a practical control interface orthogonal to re-training or task-specific fine-tuning.

This paper studies emotion controllability in LALMs from a neuron-level perspective, focusing on instruction-following EVC. Our central hypothesis is that compact neuron subsets associated with emotional expression exist in speech-generative LALMs, and that intervening on these subsets can causally affect measured EVC outcomes.
We center the study around the following research questions:
\begin{enumerate}
    \item \textbf{Existence.} Do ESNs exist in speech-generative LALMs, and does intervening on them causally affect EVC outcomes? If so, do these effects primarily alter emotional expressiveness or linguistic content?
    \item \textbf{Identification.} Which neuron selection criteria most effectively isolate ESNs for controllable EVC, and how do mask sparsity and identification data size influence results? 
    \item \textbf{Inference-Time Controllability.} How do intervention type and strength affect emotion control and content preservation in EVC?
    \item \textbf{Localization.} Where do ESNs concentrate across modules and across layers in LALMs, and what does this imply for controllable EVC?
\end{enumerate}
To answer these questions, we adopt a four-stage activation-based pipeline: (1) activation sampling during EVC with hooks on decoder MLP gates; (2)  filtering successful EVC instances for activation aggregation; 
(3) ESN identification by various neuron ranking criteria: frequency-based, entropy-based, mean-deviation-based, and contrastive-margin-based; and (4) inference-time control via targeted steering, additive injection, clamping, and deactivation, keeping model weights untouched.

\begin{figure*}[t]
    \centering
    \includegraphics[width=0.9\textwidth]{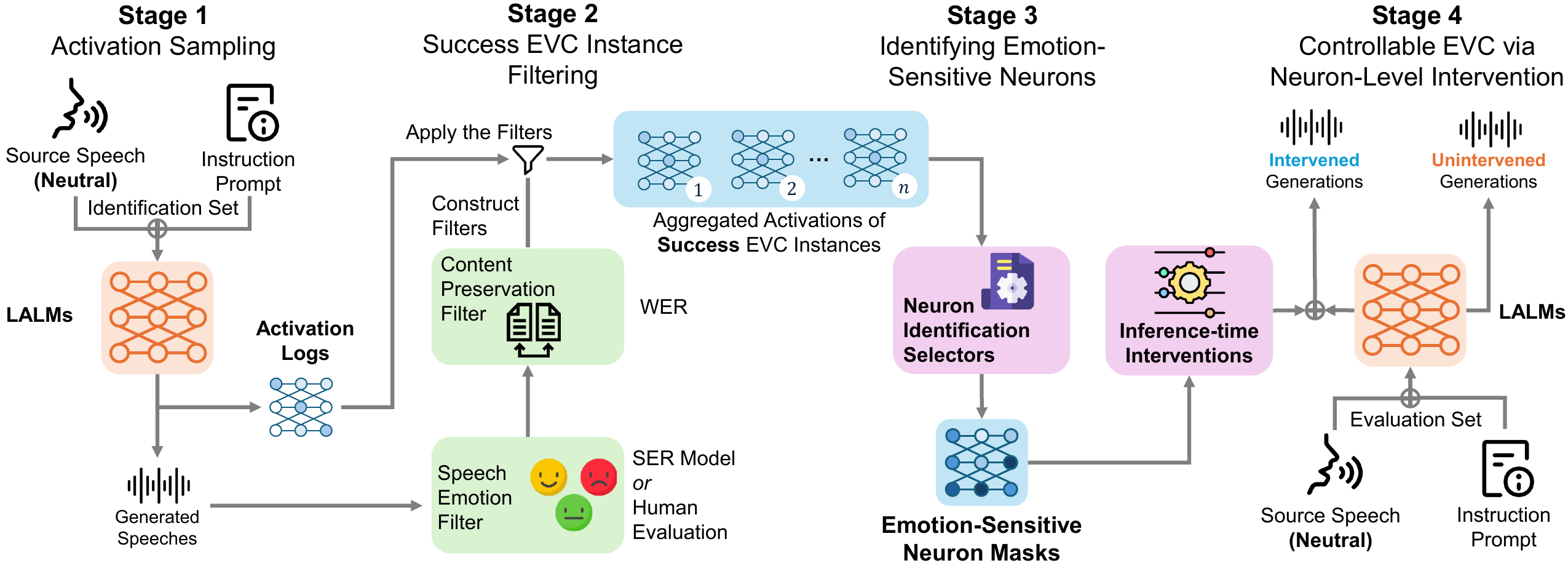}
    \caption{The overview of our four-stage pipeline for identifying and manipulating emotion-sensitive neurons in LALMs for EVC.}
    \label{fig:pipeline}
\end{figure*}

However, EVC's inherently multi-objective nature makes isolating qualified conversions complex \cite{10.1016/j.specom.2021.11.006}. This motivates a multi-stage filtering and identification protocol (Figure~\ref{fig:pipeline}) that aggregates activations only from outputs that achieve both emotion realization and content preservation. We later adapt the same two axes used for evaluating intervention effects. 

Using the Emotional Speech Dataset (ESD) \cite{10.1016/j.specom.2021.11.006}, we evaluate three open speech-generative LALMs: Qwen2.5-Omni-7B \cite{xu2025qwen25omnitechnicalreport}, MiniCPM-o 4.5 \cite{yao2024MiniCPMvgpt4vlevelmllm}, and Kimi-Audio \cite{kimiteam2025kimiaudiotechnicalreport}. Even without intervention, instruction-following EVC proves challenging: baseline emotion match rates are modest, target emotions vary substantially in difficulty, and stronger affect expression does not necessarily coincide with better content preservation. Despite this difficult baseline, neuron-level intervention produces consistent and interpretable effects. Compact ESN masks induce structured self–cross patterns rather than arbitrary degradation, indicating that they causally influence emotional rendering.

Controllability first depends on the identification-time selection criteria. Our results show that the type of evidence emphasized by a selector directly impacts ESN effectiveness for emotion control. In particular, contrastive-margin- and mean-deviation-based selectors isolate substantially more effective ESNs than frequency- or entropy-based ones. Masks that are too sparse fail to capture the distributed emotion signal, whereas overly large masks admit shared or generic generation features and reduce specificity. A similar trade-off holds for identification data: ESN estimates stabilize with a relatively small pool of qualified conversions, after which additional examples yield diminishing returns and may increase cross-emotion interference.

Intervention design then shapes inference-time controllability. 
In our Qwen2.5-Omni-7B strength sweep, moderate steering strengths yield the best observed balance between target-emotion gain and content preservation, whereas stronger interventions dramatically degrade linguistic fidelity, indicating that effective EVC control depends on calibrated rather than brute-force manipulation. These gains are observed in both automatic evaluation and human listening, where intervened samples are preferred in 62\% of pairwise comparisons on average, supporting that ESN-masked models generate more perceptible emotion shifts. Localization analyses further show that the most actionable ESNs concentrate in intermediate-to-late decoder MLP layers on the language-model side across all three LALMs, while analogous interventions in downstream synthesis MLPs are largely inert. This is consistent with the view that emotion-related control is more accessible on the LM side before acoustic rendering than in the late-stage synthesis modules we tested.

\textbf{Contributions.} This paper provides, to our knowledge, one of the first systematic neuron-level analyses of emotion control in speech-generative LALMs under an EVC setting. We show that compact ESN masks are causally actionable, enabling training-free intervention that produces consistent target-emotion control beyond purely correlational analysis. Within a unified framework, we systematically compare ESN selectors, mask sparsity, success set size, and inference-time intervention designs, identifying contrastive-margin- and mean-deviation-based criteria as the most favorable among the tested selectors for isolating actionable ESNs. We further localize the most impactful emotion-control signals to intermediate-to-late decoder MLP layers on the language-model side, while analogous interventions in the downstream synthesis stack are largely ineffective. Finally, we show that these effects generalize beyond the identification split and are corroborated by human listening, establishing a practical and mechanistically grounded route to training-free emotion control in speech-generative LALMs.

\section{Related Work}
Affective speech synthesis and conversion aim to generate or transform speech while controlling paralinguistic affect, most prominently emotion \cite{10065433}. In emotional speech synthesis, controllability is typically achieved by learning explicit style or emotion representations and conditioning generation, such as via reference encoders \cite{skerry2018towards} or global style tokens \cite{wang2018style}, with subsequent extensions incorporating semi- or weak supervision \cite{wu2019end, liu2021expressive, lei2022msemotts, zhang2023iemotts}. Complementary work explores continuous affect control, including emotion intensity prediction or transfer \cite{lorenzo2018investigating, lei2021fine, schnell21_ssw, cho24_interspeech}, while recent approaches investigate token-level manipulation for expressive text-to-speech (TTS) \cite{xie2025emosteerttsfinegrainedtrainingfreeemotioncontrollable}.  EVC \cite{10.1016/j.specom.2021.11.006} further studies mapping between emotional states, including non-parallel adversarial and multi-domain frameworks \cite{cao20b_interspeech, he21b_interspeech, 9413391, 10.1016/j.specom.2022.09.002}, diffusion-based approaches \cite{prabhu23_emoconvdiff}, and representation-factorized models \cite{kreuk-etal-2022-textless}.

In parallel, neuron-level interpretability work
that localizes functional units in deep neural networks \cite{bau2020understanding, bau2019identifying, singla2022audio} has encouraged explorations on neuron/activation-level control in LLMs \cite{yu-ananiadou-2024-neuron, turner2024steeringlanguagemodelsactivation, rimsky-etal-2024-steering, silva-etal-2025-steering}, alongside emerging interests for LLM-centric multimodal models \cite{wu2024andaudionetworkdissection, fang2024towards, huo-etal-2024-mmneuron, yang2025audiolenscloserlookauditory}.
Among them, audio-language models capable of speech understanding and generation \cite{yao2024MiniCPMvgpt4vlevelmllm, xu2025qwen25omnitechnicalreport, kimiteam2025kimiaudiotechnicalreport} enable studying affective control in instruction-following speech generation settings. Recent studies report affective circuits in text LLMs \cite{lee-etal-2025-large, wang2025llmsfeelemotioncircuits}, however, there have been very limited attempts to utilize emotion-related functional units in LALMs \cite{zhao2026discoveringcausallyvalidatingemotionsensitive}.
Unlike prior work, which primarily targets recognition, representation learning, or token-level manipulation, we probe the synthesis pathway in speech-generative LALMs by logging activations during EVC generation that yield successful conversions and applying training-free neuron-level interventions to directly steer emotion expressiveness.

\section{Methodology}
We study emotion control in speech-generative LALMs through a neuron-level intervention framework tailored to instruction-following EVC. While activation-based interpretability and steering methods have been explored in text and multimodal models, applying them to speech generation is non-trivial.
Unlike text classification probes, speech generation exhibits two key challenges: (1) generation is harmed by diverse failure modes (hallucinated content, silence, refusals) that stem from language-model (LM) components; and (2) task success is inherently multi-objective, requiring both \emph{emotional style} transfer and \emph{linguistic content} preservation. To address these challenges, we design a four-stage pipeline that explicitly filters qualified EVC instances before activation aggregation (Figure~\ref{fig:pipeline}). This structured protocol enables more reliable identification of candidate emotion-sensitive neurons and interventional testing of their functional relevance in speech-generative LALMs.

\subsection{Activation Sampling}
\label{subsec:recording}

For each source speech utterance $x$ in the identification set, we instruct the LALM to perform EVC and generate audio $y$ of the target emotion $e_{tgt}$. The instruction prompt is fixed across conditions, and we record activations during the forward passes of the modules that mediate generation. 

\textbf{Hooking Position.}
We instrument the decoder-side feed-forward blocks (MLPs) because they offer a stable, high-bandwidth location for neuron-level interventions \cite{bau2019identifying, turner2024steeringlanguagemodelsactivation}. 
Many LALMs employ gated MLPs (e.g., SwiGLU variants) \cite{shazeer2020glu}, which offer a natural position for hooking. Concretely, let $u_{l,t}, v_{l,t}\in\mathbb{R}^{D_l}$ be the two pre-activation streams at layer $l$ and decoding step/token position $t$, and let
\begin{equation}
g_{l,t} = \phi(u_{l,t}) \in \mathbb{R}^{D_l}
\end{equation}
denote the gated branch after the nonlinearity $\phi(\cdot)$ (e.g., $\mathrm{SiLU}$). The MLP output is $h_{l,t}=W_o\left(g_{l,t}\odot v_{l,t}\right)$, where $\odot$ is elementwise product. We log the scalar neuron activations
$a_{l,n,t} = [g_{l,t}]_n,$
for neuron index $n\in\{1,\dots,D_l\}$. These logged values serve as the basis for our identification statistics and intervention hooks.

In addition to the MLPs in the language-model component of LALMs, we also examine whether the synthesis MLP layers also contain ESNs, which we analyze later in \S~\ref{sec:location}. %
For each conversion attempt, we store per-neuron counts and moments computed over the full activation tensor produced during generation; we do not apply any success-based selection or modality restriction at this step.

\subsection{Success Filtering and Set Construction}
\label{subsec:filtering}

A key design choice is to record and utilize activations only on conversions that plausibly satisfy EVC. Raw generations include diverse failures, as we showed in Figure~\ref{fig:example}, and mixing them into activation statistics can obscure emotion-related signals. For each generated audio $y$, we apply a two-axis filter aligned with the standard decomposition of voice conversion evaluation into \emph{speaking style} and \emph{linguistic content} \cite{10.1016/j.specom.2021.11.006}.

\textbf{Emotion check via held-out judge.}
We employ a \emph{held-out} SER model as an automatic judge. It predicts an emotion label $\hat{e}(y)$. We accept the generation as emotion-matched if
$\hat{e}(y) = e_{tgt}$.
In a subset of experiments, we also conduct human listening experiments to validate the automatic judges' reliability; \S~\ref{sec:human} discusses differences and agreement patterns.

\textbf{Content preservation via word error rate (WER) threshold.}
We examine linguistic content preservation by comparing the decoded text content produced by the LALM with the source transcription of $x$ (or its reference transcript). 
Importantly, since the LALM emits decoded text alongside audio, we use the decoded text directly rather than transcribing $y$ with a separate automatic speech recognition (ASR) engine for evaluation of preservation. We set a fixed threshold $\tau_{\text{wer}} = 0.15$ to define/filter the successful instance across our experiments.

\textbf{Success instance sets.}
For each target emotion $e\in\mathcal{E}$, we denote the filtered success set as $\mathcal{S}_e$. 
The size of this set is capped by a parameter $c$ (the set is uniformly sampled if $|\mathcal{S}_e|>c$) to balance emotions (later analyzed in \S~\ref{success_size}).

\subsection{Activation Aggregation and Emotion-Sensitive Neuron Identification}
\label{subsec:identification}
\subsubsection{Activation Statistics on Filtered Successes}
\label{sec:selectors}
Using only the filtered success set $\mathcal{S}_e$, our method computes neuron-wise activation \emph{probabilities} from positive-activation counts. For each neuron $(l,n)$ and emotion $e$, we accumulate
\begin{align}
K^{(e)}_{l,n}
&\leftarrow K^{(e)}_{l,n}
+ \sum_{(x,y)\in\mathcal{S}_e}\sum_t\,
\mathbb{I}(a^{(x\rightarrow y)}_{l,n,t} > 0), \\
T_e
&\leftarrow T_e
+ \sum_{(x,y)\in\mathcal{S}_e}\sum_t 1.
\end{align}
We then define the activation probability
$P^{(e)}_{l,n}=\frac{K^{(e)}_{l,n}}{T_e}.$
Thus, $P^{(e)}_{l,n}$ measures how often neuron $(l,n)$ is active on successful EVC instances of target emotion $e$.

Let $N = \sum_{l=1}^{L} D_l$ denote the total number of neurons across all instrumented layers, and let $r\in(0,1)$ be the selection rate. In all four methods below, the implementation selects
$N_{\text{sel}}=\lfloor rN \rfloor$
neurons per emotion.

\subsubsection{Neuron Ranking Methods}
\begin{itemize}[leftmargin=*, itemsep=3pt]
    \item \textbf{LAP (Activation Probability)} \cite{cunningham2023sparse, gurnee2024universal, voita-etal-2024-neurons}.  
    For each emotion $e$, neurons are ranked directly by their activation probability, i.e., $\mathrm{LAP}^{(e)}_{l,n}=P^{(e)}_{l,n}$, and the top $N_{\text{sel}}$ are selected.

    \item \textbf{LAPE (Activation Probability Entropy)} \cite{tang-etal-2024-language, huo-etal-2024-mmneuron, namazifard-poech-2025-isolating}.  
    This implementation applies a two-stage procedure. First, for each neuron, activation probabilities are normalized across emotions, $\tilde P^{(e)}_{l,n}=P^{(e)}_{l,n}/\sum_{e'}P^{(e')}_{l,n}$, and an entropy score $H_{l,n}=-\sum_e \tilde P^{(e)}_{l,n}\log \tilde P^{(e)}_{l,n}$ is computed, where lower entropy indicates stronger emotion selectivity. Neurons that are globally too inactive are then removed using a probability threshold $\tau_p$ computed from a global percentile over all $P^{(e)}_{l,n}$ values; a neuron is kept only if it exceeds $\tau_p$ for at least one emotion. Among the remaining neurons, the lowest-entropy $N_{\mathrm{cand}}=\lfloor \min(1,5r)N \rfloor$ form a candidate pool. Finally, for each target emotion $e$, the method selects the top $N_{\text{sel}}$ neurons from this pool by $P^{(e)}_{l,n}$.

    \item \textbf{MAD (Mean Activation Deviation)} \cite{bau2018identifying, 10.1609/aaai.v33i01.33016309, mean-centring}.  
    For each neuron, the implementation first computes its mean activation probability across emotions, $\bar P_{l,n}=\frac{1}{|\mathcal{E}|}\sum_{e'}P^{(e')}_{l,n}$. For target emotion $e$, the score is then the absolute deviation from this mean:
    \begin{equation}
    \mathrm{MAD}^{(e)}_{l,n} = \left|P^{(e)}_{l,n} - \bar P_{l,n}\right|.
    \end{equation}
    Hence, this implementation favors neurons whose activation probability for emotion $e$ differs strongly, in either direction, from their average behavior across emotions.

    \item \textbf{CAS (Contrastive Activation Selection)} \cite{zhao2025findingculturesensitiveneuronsvisionlanguage, zhao2026discoveringcausallyvalidatingemotionsensitive}.  
    For each neuron, let $P^{(1)}_{l,n}$ and $P^{(2)}_{l,n}$ denote its largest and second-largest activation probabilities across emotions, respectively, and let $e^{(1)}_{l,n}$ be the corresponding top emotion. The margin is $\Delta_{l,n}=P^{(1)}_{l,n}-P^{(2)}_{l,n}$. For target emotion $e$, the method ranks a neuron by this margin only when $e=e^{(1)}_{l,n}$; otherwise, its score is set to $-\infty$. It then selects the top $N_{\text{sel}}$ neurons for each emotion.

    \item \textbf{Random baseline.} 
    As a control, we additionally sample $N_{\text{sel}}=\lfloor rN\rfloor$ neurons uniformly at random without replacement from all $N=L\times D$ hooked neurons, and map them back to their layers to form a mask with the same sparsity as the other selectors.
\end{itemize}

\subsection{Inference-Time Emotion Control via Neuron-Level Interventions}
\label{subsec:intervention}

To test whether the identified neurons are causal for EVC, we intervene on the post-activation gate values of the hooked MLPs during generation, while keeping all model weights fixed. Concretely, for a hooked layer $l$, let $g_{l,t}\in\mathbb{R}^{D_l}$ denote the activated gate vector at generation step $t$, and let $\mathcal{I}^{(m,e)}_l$ be the set of selected neuron indices for method $m$ and target emotion $e$. Interventions are applied only on dimensions in $\mathcal{I}^{(m,e)}_l$, before the elementwise product with the parallel MLP branch and the output projection.
\begin{itemize}[leftmargin=*, itemsep=3pt]
    \item \textbf{Targeted steering (gain scaling)} \cite{turner2024steeringlanguagemodelsactivation}.
For selected neurons, we multiply the activated gate values by a constant gain $1+\alpha$ ($\alpha \ge 0$). In compact form,
\begin{equation}
[\tilde g_{l,t}]_n =
\begin{cases}
(1+\alpha)[g_{l,t}]_n, & n\in \mathcal{I}^{(m,e)}_l,\\
[g_{l,t}]_n, & \text{otherwise.}
\end{cases}
\end{equation}

    \item \textbf{Additive shift} \cite{rimsky-etal-2024-steering}.
As an alternative intervention, we add a constant offset $\alpha$ ($\alpha \ge 0$) to the selected gate dimensions:
\begin{equation}
[\tilde g_{l,t}]_n =
\begin{cases}
[g_{l,t}]_n + \alpha, & n\in \mathcal{I}^{(m,e)}_l,\\
[g_{l,t}]_n, & \text{otherwise.}
\end{cases}
\end{equation}
Unlike steering, this is a uniform scalar shift and does not use neuron-specific reference values.

    \item \textbf{Floor clamping} \cite{templeton2024scaling}.
We also test a minimum-activation constraint, in which each selected dimension is clamped from below by a scalar threshold $\alpha$ ($\alpha \ge 0$):
\begin{equation}
[\tilde g_{l,t}]_n =
\begin{cases}
\max\left([g_{l,t}]_n,\alpha\right), & n\in \mathcal{I}^{(m,e)}_l,\\
[g_{l,t}]_n, & \text{otherwise.}
\end{cases}
\end{equation}
This ensures that selected neurons remain at or above a fixed activation floor during generation.

    \item \textbf{Deactivation.}
Finally, we test necessity by zeroing the selected activated gate dimensions:
\begin{equation}
[\tilde g_{l,t}]_n =
\begin{cases}
0, & n\in \mathcal{I}^{(m,e)}_l,\\
[g_{l,t}]_n, & \text{otherwise.}
\end{cases}
\end{equation}
This removes the contribution of the selected neurons while leaving the rest of the computation unchanged.
\end{itemize}

\subsection{Evaluation Metrics and Protocol}

Echoing the filtering protocol in \S~\ref{subsec:filtering} and in accordance with the EVC objectives, we evaluate three aspects: \textbf{emotion shift}, \textbf{content preservation}, and \textbf{naturalness}. All generations are resampled to 16 kHz.

\begin{itemize}
    \item \textbf{Emotion shift success.}
    We report \emph{Emotion match rate}, the fraction of converted samples whose predicted emotion matches $e_{\text{tgt}}$. (1) SER judges are provided with $(x, \hat{x}_{e_{\text{tgt}}})$, for \textbf{emotion2vec+large} \cite{ma-etal-2024-emotion2vec}, we directly use the predicted label. For \textbf{Qwen3-Omni-30B} \cite{xu2025qwen3omnitechnicalreport}, we prompt the model to perform five-way multiple-choice emotion classification.
    (2) \textbf{Human evaluation} uses pairwise A/B preference. See details in \S~\ref{sec:human}.
    \item \textbf{Content preservation.}
    We compute WER between the reference transcript and the extracted decoded text from the LALMs, normalized by the whisper-normalizer.\footnote{\url{https://pypi.org/project/whisper-normalizer/}}

    \item \textbf{Naturalness.}
    We estimate perceptual quality using the UTokyo-SaruLab Mean Opinion Score (UTMOS) \cite{saeki22c_interspeech}, a non-intrusive mean opinion score (MOS) estimator. 
\end{itemize}

\textbf{Interventions and evaluation conditions.}
For each emotion $e_{\text{mask}}\in E$, we identify a set of ESNs using each selector on the \textsc{Identification} split. We then evaluate emotion sensitivity on held-out utterances by comparing generation with and without intervention under two complementary conditions: (1) \textbf{self-effect} ($e_{\text{mask}} {=} e_{\text{tgt}}$), where the model is instructed to generate emotion $e_{\text{tgt}}$ and is intervened with the ESN mask identified for that same emotion; and (2) \textbf{cross-effect} ($e_{\text{mask}} {\neq} e_{\text{tgt}}$), where the model is instructed to generate emotion $e_{\text{tgt}}$ but is intervened with an ESN mask identified for a different emotion. For positive interventions (scaling, additive, and clamping), the expected effect is to strengthen evidence associated with $e_{\text{mask}}$. Comparing self- and cross-effect therefore helps distinguish emotion-specific modulation from broad degradation or nonspecific perturbation.

\section{Experiment Setup}

\subsection{Models}
As outlined, we evaluate three LALMs: \textbf{Qwen2.5-Omni-7B} \cite{xu2025qwen25omnitechnicalreport}, \textbf{Kimi-Audio} \cite{kimiteam2025kimiaudiotechnicalreport}, \textbf{MiniCPM-o 4.5} \cite{yao2024MiniCPMvgpt4vlevelmllm}, all of which are multimodal models with audio-generation capability. They were selected for their strong benchmark performance and instruction-following ability. Their architectural differences further enable comparative interpretability analysis. For automated conversion-success judgment, we employ two models: \textbf{emotion2vec+large} \cite{ma-etal-2024-emotion2vec} serves as a speech emotion recognition (SER) judge from a primarily acoustic perspective. On a 500-sample (100 per emotion) held-out subset of ESD, emotion2vec+large achieves 97.6\% overall SER accuracy. In addition, we purposefully introduce an advanced LALM, \textbf{Qwen3-Omni-30B} \cite{xu2025qwen3omnitechnicalreport} as a supplementary LALM-based judge to assess potential bias toward linguistic content in emotion evaluation.

\subsection{Dataset}
Our primary test bed is the \textbf{Emotional Speech Database (ESD)} \cite{10.1016/j.specom.2021.11.006}, which contains 20 speakers, each with 350 parallel utterances per emotion across five categories (``neutral'', ``happy'', ``angry'', ``sad'', ``surprise''). To avoid language-dependent confounds, we restrict the dataset to the 10 English speakers. We further split the data by utterance index (parallel across emotions) to prevent lexical leakage across emotions. As summarized in Table~\ref{tab:speaker_split}, we use a hybrid split: a 6/2/2 speaker partition, with utterance-index splits within the seen-speaker portion. Unless noted otherwise, all reported results are computed on \textsc{Test-Seen} and \textsc{Test-Unseen}, with no overlap in speakers or utterance indices relative to the identification or development splits.
\begin{table}[ht]
\caption{\textbf{Speaker and utterance-ID split of ESD.} Utterance IDs are parallel across the five emotions for each speaker.}
\label{tab:speaker_split}
\centering
\resizebox{\columnwidth}{!}{%
\begin{tabular}{@{}lccc|r@{}}
\toprule
\textbf{Speakers}       & S1--S6         & S7--S8         & S9--S10        & \textbf{Total} \\ \midrule
\textsc{Identification} & 001--200 (200) & --             & --             & 1200           \\
\textsc{Development}    & 201--250 (50)  & 001--250 (250) & --             & 1100           \\
\textsc{Test-seen}      & 251--350 (100) & 251--350 (100) & --             & 800            \\
\textsc{Test-unseen}    & --             & --             & 001--350 (350) & 700            \\ \bottomrule
\end{tabular}%
}
\vspace{-3mm}
\end{table}

\subsection{Task Implementation}
Although EVC is inherently bidirectional, we focus on converting neutral speech to a target emotion $e_{tgt} \in \{\text{happy, angry, sad, surprise}\}$.
For LALM-based conversion, we use a fixed, emotion-agnostic system prompt instructing the model to convert speech emotion while preserving content, together with a user prompt that provides (1) the input audio and (2) the EVC instruction that specifies a target emotion (Figure~\ref{fig:example} shows an example). We employ deterministic decoding when available (greedy, temperature $=0$) to facilitate reproducibility.

\section{Results}
\subsection{Baseline Performance}
\begin{table}[ht]
\caption{\textbf{ Baseline EVC performance} of the three evaluated LALMs on ESD, before any neuron-level intervention. We report the three EVC axes used throughout the paper: target-emotion match rate in \% (emotion2vec / Qwen3), content preservation (WER in \%), and naturalness (UTMOS) on its 1–5 scale (the higher the more natural).}
\label{tab:baseline}
\centering
\resizebox{0.9\columnwidth}{!}{%
\begin{tabular}{@{}lccc@{}}
\toprule
         & Qwen2.5-Omni-7B                 & MiniCPM-o 4.5 & Kimi-Audio                      \\ \midrule
Angry    & 0.13 / 10.93                    & 3.73 / 12.93  & \textbf{10.60} / \textbf{16.00} \\
Happy    & \textbf{45.40} / \textbf{10.13} & 2.13 / 5.33   & 39.60 / 7.87                    \\
Sad      & \textbf{52.20} / 21.60          & 2.33 / 9.80   & 42.89 / \textbf{31.93}          \\
Surprise & 2.67 / 5.07                     & 0.07 / 3.47   & \textbf{7.13} / \textbf{6.67}   \\ \midrule
Avg.     & \textbf{25.10} / \textbf{11.93} & 2.07 / 7.88   & 15.62 / 7.13                    \\ \midrule
WER      & 19.00                           & 2.95          & \textbf{2.82}                   \\
UTMOS & 4.00                            & \textbf{4.26} & 3.18                            \\ \bottomrule
\end{tabular}%
}
\vspace{-3mm}
\end{table}

\begin{figure*}[t]
    \centering

    \begin{minipage}[t]{0.66\linewidth}
        \vspace{0pt}
        \centering
        \captionof{table}{\textbf{Comparison of ESN identification methods under activation steering, reported relative to the unintervened baseline in Table~\ref{tab:baseline}.}}
        \label{tab:main}
        \resizebox{\linewidth}{!}{%
        \begin{tabular}{@{}llccccc@{}}
        \toprule
                      &          & \multicolumn{3}{c}{$\Delta$Emotion Match vs. Base (emotion2vec / Qwen3)}             &                             &                             \\ \cmidrule(lr){3-5}
        LALM          & Selector & Self-Effect $\uparrow$ & Cross-Effect Avg. $\downarrow$ & Self--Cross Gap $\uparrow$ & WER $\downarrow$ ($\Delta$) & UTMOS $\uparrow$ ($\Delta$) \\ \midrule
                      & Random   & +0.52 / +0.01          & --                             & --                         & 22.54 (+3.54)               & 3.99 (-0.01)                \\
                      & LAP      & -2.15 / +0.04          & -1.89 / +0.07                  & -0.25 / -0.03              & \textbf{18.94 (-0.06)}      & 3.92 (-0.08)                \\
        Qwen2.5-      & LAPE     & +1.82 / +1.07          & +2.21 / +1.83                  & -0.39 / -0.77              & 34.70 (+15.70)              & 3.94 (-0.06)                \\
        Omni-7B       & MAD      & +2.62 / +2.82          & +0.81 / +0.05                  & +1.81 / +2.30              & 24.22 (+5.22)               & \textbf{3.99 (-0.01)}       \\
                      & CAS      & \textbf{+4.27 / +3.60} & \textbf{+0.34 / -0.05}         & \textbf{+3.93 / +3.65}     & 29.65 (+10.65)              & \textbf{3.99 (-0.01)}       \\
                      & CAS-H    & +3.15 / +2.92          & +0.62 / +0.11                  & +2.53 / +2.81              & 24.56 (+5.56)               & \textbf{3.99 (-0.01)}       \\ \midrule
                      & Random   & -0.05 / +0.14          & --                             & --                         & 3.14 (+0.19)                & 4.25 (-0.01)                \\
                      & LAP      & -0.33 / -0.38          & -0.42 / -0.16                  & \textbf{+0.09} / -0.23     & 8.98 (+6.03)                & 4.25 (-0.01)                \\
        MiniCPM-o 4.5 & LAPE     & -0.28 / +0.33          & -0.18 / +0.34                  & -0.10 / -0.01              & \textbf{2.81 (-0.14)}       & 4.25 (-0.01)                \\
                      & MAD      & -0.22 / +0.33          & -0.17 / +0.16                  & -0.05 / -0.18              & 2.84 (-0.11)                & \textbf{4.26 (0.00)}        \\
                      & CAS      & \textbf{-0.20 / +0.40} & \textbf{-0.08 / +0.09}         & -0.12 / \textbf{+0.31}     & 2.95 (0.00)                 & 4.25 (-0.01)                \\ \midrule
                      & Random   & -0.76 / -0.12          & --                             & --                         & 3.72 (+0.90)                & \textbf{3.17 (-0.01)}       \\
                      & LAP      & \textbf{+1.52} / +2.71 & +0.35 / +1.49                  & +1.17 / +1.22              & 4.13 (+1.31)                & 3.15 (-0.03)                \\
        Kimi-Audio    & LAPE     & -0.72 / -0.48          & -0.89 / \textbf{+0.46}         & +0.16 / -0.94              & \textbf{2.90 (+0.08)}       & 3.16 (-0.02)                \\
                      & MAD      & -0.90 / +1.17          & \textbf{-0.02} / +1.15         & -0.88 / +0.02              & 4.69 (+1.87)                & 3.15 (-0.03)                \\
                      & CAS      & +1.19 / \textbf{+7.65} & -0.19 / +0.96                  & \textbf{+1.38 / +6.70}     & 4.74 (+1.92)                & 3.16 (-0.02)                \\ \bottomrule
        \end{tabular}}
    \end{minipage}\hfill
    \begin{minipage}[t]{0.31\linewidth}
        \vspace{0pt}
        \centering
        \begin{subfigure}[b]{0.5\linewidth}
            \centering
            \includegraphics[width=\linewidth]{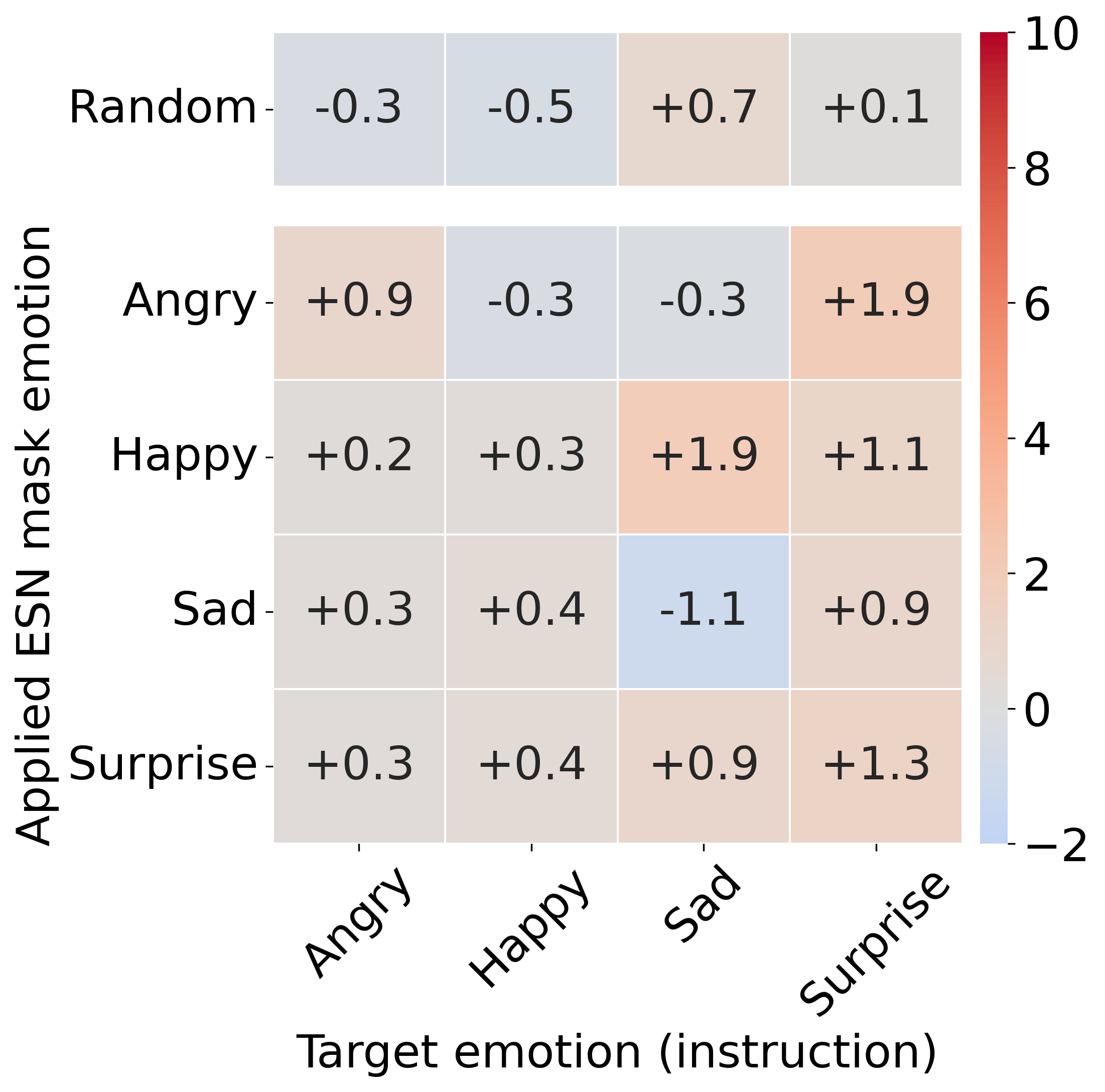}
            \caption{\textbf{LAP}}
        \end{subfigure}\hfill
        \begin{subfigure}[b]{0.5\linewidth}
            \centering
            \includegraphics[width=\linewidth]{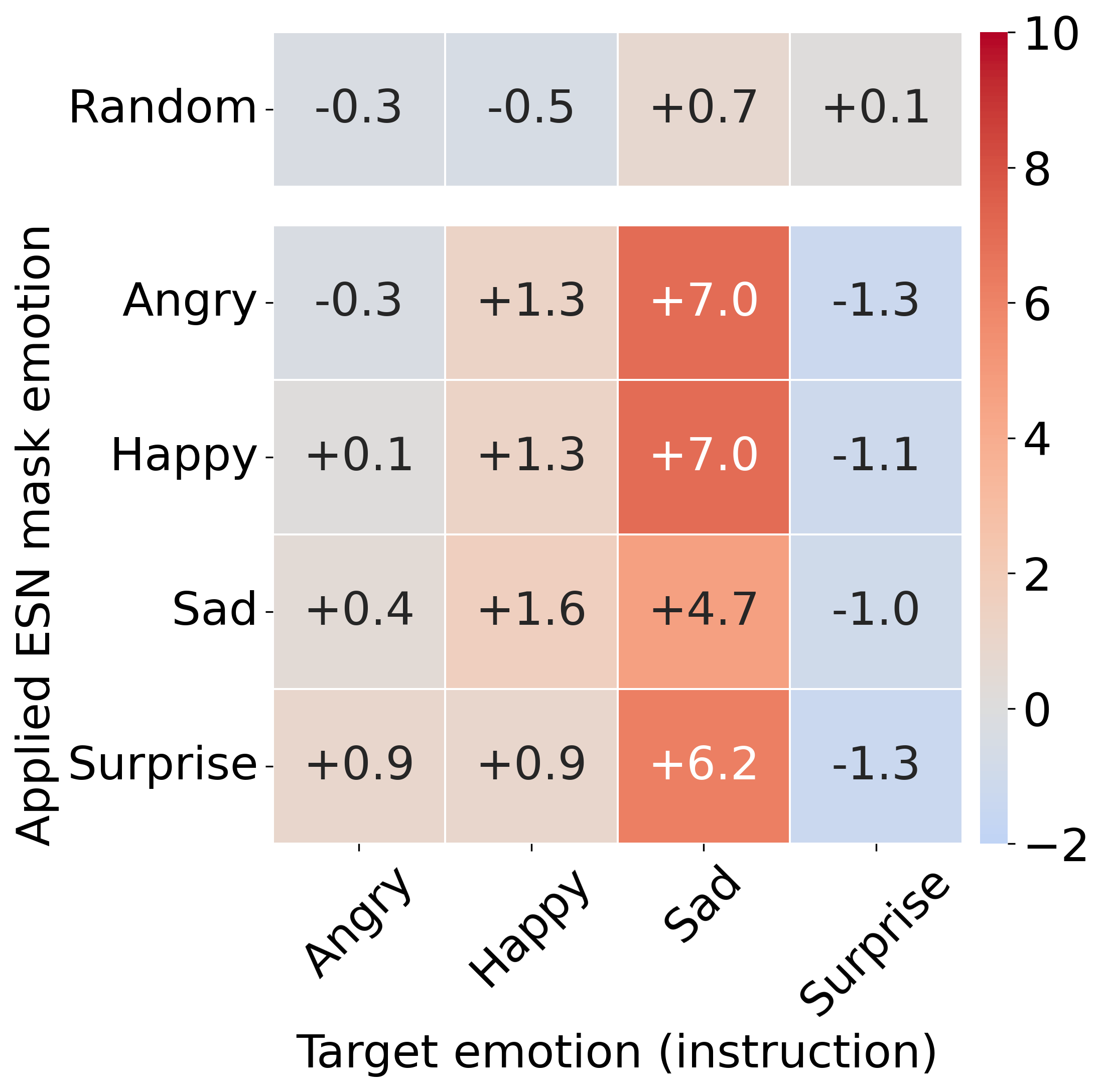}
            \caption{\textbf{LAPE}}
        \end{subfigure}\hfill

        \vspace{2mm}

        \begin{subfigure}[b]{0.5\linewidth}
            \centering
            \includegraphics[width=\linewidth]{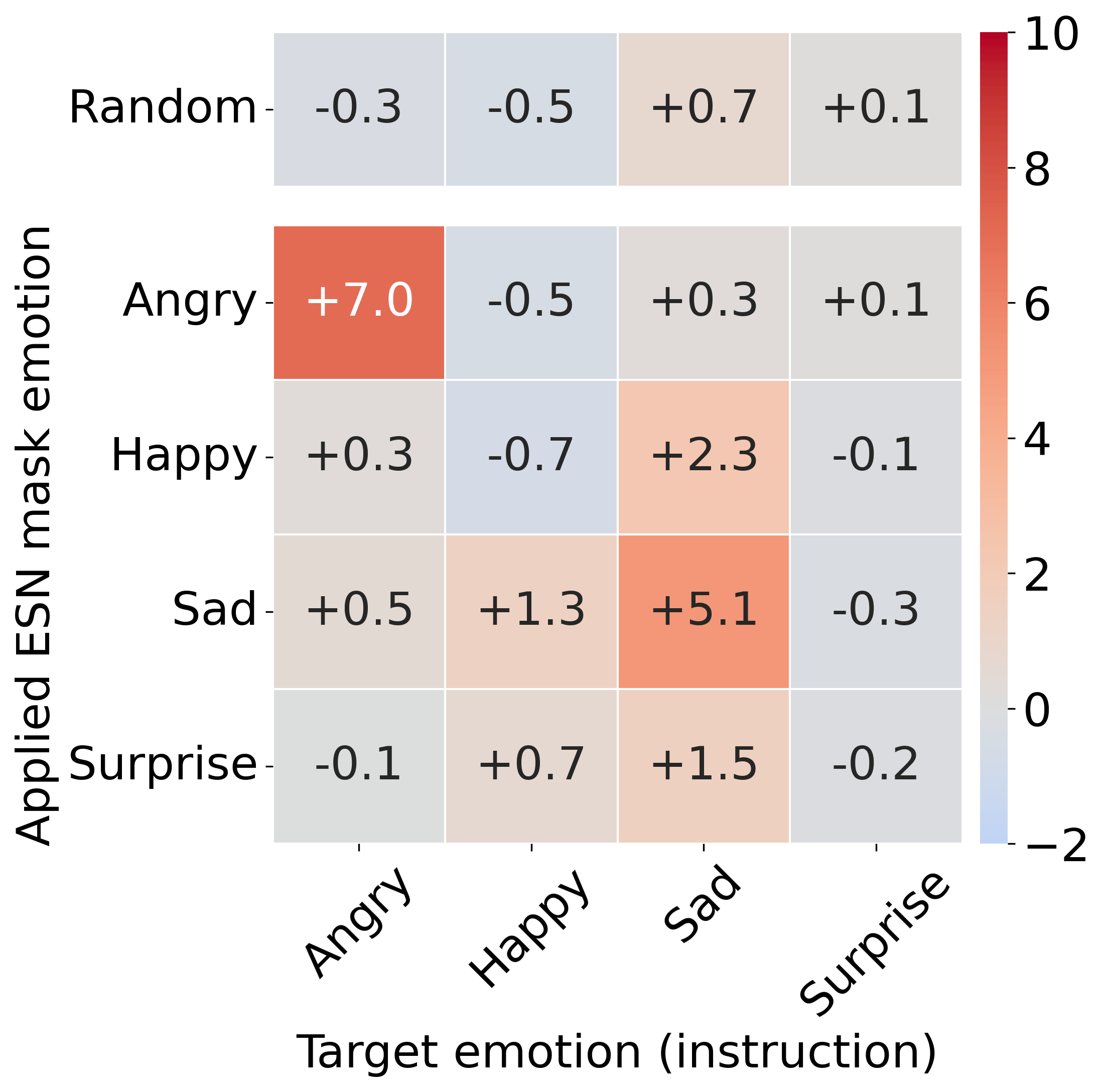}
            \caption{\textbf{MAD}}
        \end{subfigure}\hfill
        \begin{subfigure}[b]{0.5\linewidth}
            \centering
            \includegraphics[width=\linewidth]{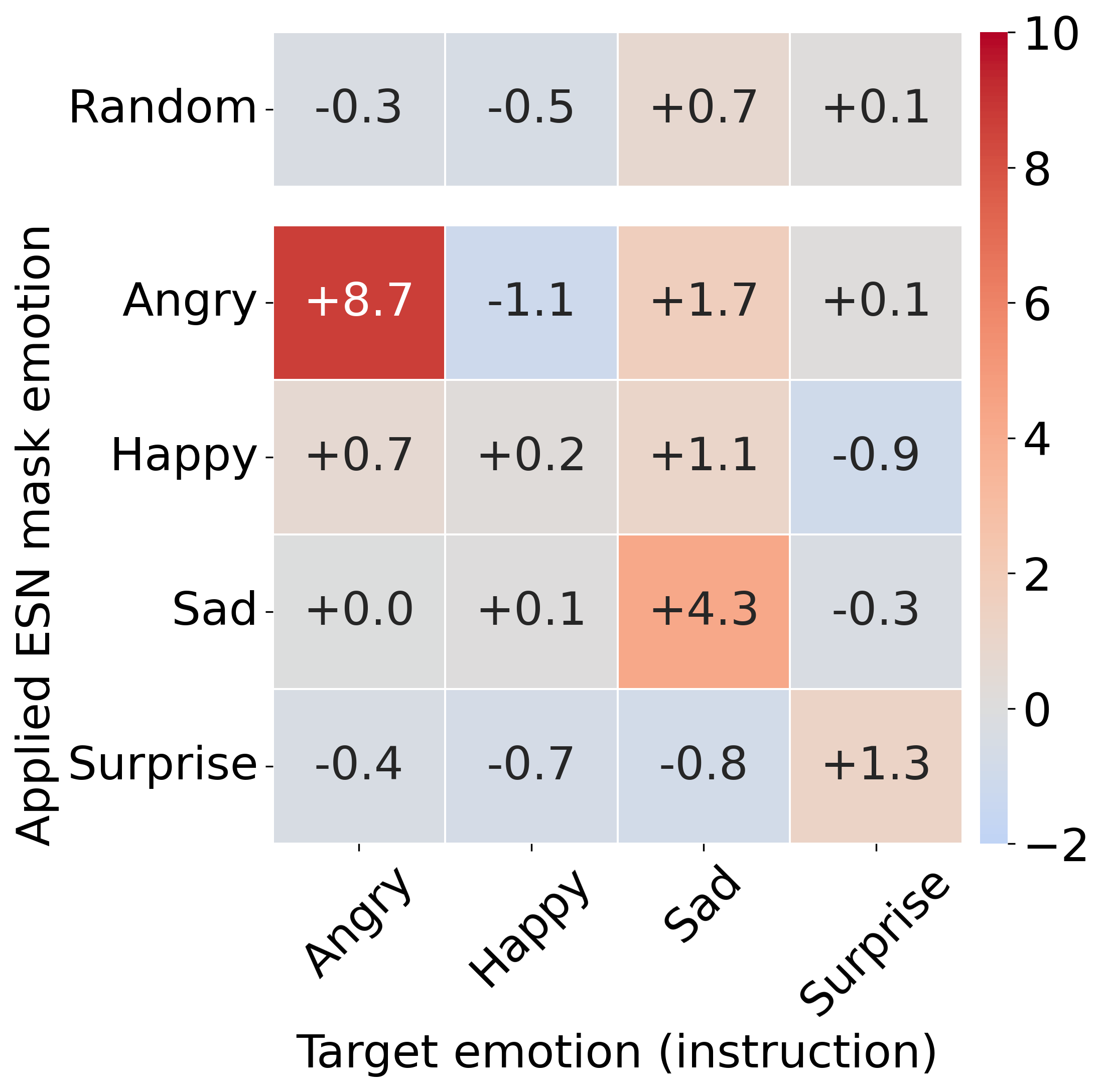}
            \caption{\textbf{CAS}}
        \end{subfigure}

    \end{minipage}

    \caption{(Left) Intervention effects on EVC, under activation steering ($c{=}50$, $\alpha{=}1.0$). For each selector, we report changes in target-emotion match \textbf{relative to the unintervened baseline in Table~\ref{tab:baseline}}, together with the corresponding changes in content preservation (WER) and post-intervention naturalness (UTMOS). The CAS-H row reports the results using human-filtered instances, see explanations in \S~\ref{sec:human}.
    (Right) \textbf{Per-emotion emotion match rate changes by ESN selectors for Qwen2.5-Omni-7B}, under the same intervention. In each heatmap, rows denote the identified ESN mask emotion ($e_{\text{mask}}$) and columns denote the instructed target emotion ($e_{\text{tgt}}$); entries are signed changes (pp) in emotion match rate relative to the unintervened baseline. Diagonal cells represent \textbf{self-effects} (where $e_{\text{mask}} {=} e_{\text{tgt}}$), and off-diagonal cells represent \textbf{cross-effects} (averaged over all $e_{\text{mask}} {\neq} e_{\text{tgt}}$ pairs).
    }
    \label{fig:steer_id_method}

    \vspace{-2mm}
\end{figure*}

Before applying neuron-level intervention, we establish the unintervened EVC baseline of the evaluated LALMs on ESD. As shown in Table~\ref{tab:baseline}, baseline emotion match rates are modest across all three models, confirming that instruction-following EVC remains challenging in our setting. Notably, neutral is frequently predicted (e.g., 57.2\% for Qwen2.5-Omni-7B), indicating a strong bias toward emotionally conservative outputs. Kimi-Audio achieves the highest average emotion match rate (15.62\%), followed by Qwen2.5-Omni-7B (11.93\%) and MiniCPM-o 4.5 (7.88\%). Across models, sad is the easiest target emotion, whereas surprise is consistently the hardest.

Content preservation shows a contrasting trend: MiniCPM-o 4.5 and Kimi-Audio yield low WER (2.95\% and 2.82\%), while Qwen2.5-Omni-7B shows substantially higher WER (19.00\%). This reveals a clear trade-off: \textbf{models that produce stronger emotional shifts do not necessarily preserve linguistic content better.} The mismatch between emotion success and content preservation motivates our two-stage success filtering (\S~\ref{subsec:filtering}): ESN identification must be based on conversions satisfying both emotion correctness and linguistic fidelity.

\subsection{Sensitivity to Identification Parameters}

\subsubsection{Identification Methods}
Building on the baseline EVC performances in Table~\ref{tab:baseline}, we next examine which selector produces ESN masks that deliver the most favorable intervention effects, ideally across all three evaluation axes.
We compare the four selectors introduced in \S~\ref{sec:selectors} under a shared parameter setting ($c{=}50$, $r{=}0.5\%$, steering, \(\alpha{=}1.0\)).
Table~\ref{tab:main} shows a clear ranking in emotion specificity. For Qwen2.5-Omni-7B, CAS and MAD produce the strongest positive self-effects together with the largest self--cross separation, whereas LAP and LAPE are notably weaker and less selective. In particular, LAPE exhibits diffuse behavior in which off-target effects can rival or exceed on-target gains. MiniCPM-o 4.5 shows much smaller absolute changes overall, suggesting that it is comparatively less responsive to this intervention design; however, the same qualitative pattern remains.
Kimi-Audio further reinforces this conclusion, but also highlights greater judge sensitivity: CAS again gives the most favorable self--cross profile, yet the effect size differs more between the two SER judge models than in the other LALMs.

We also notice that stronger emotion gains often come with some cost in content preservation, especially for Qwen2.5-Omni-7B, where CAS and MAD increase WER more than weaker selectors. By contrast, naturalness remains remarkably stable across methods. This highlights a key finding, that the \textbf{main trade-off is not naturalness collapse but semantic drift}. Across models and selectors, UTMOS shifts are generally small across settings,  suggesting that interventions rarely cause the acoustic gibberish typical of failed conventional VC; instead, the LALM outputs remain largely intelligible and natural-sounding, but may deviate in meaning from the instructed content. 

Meanwhile, WER can increase (especially for Qwen2.5-Omni-7B under CAS/MAD), which we interpret as reflecting semantic shift rather than degradation of basic speech formation. In this sense, selector comparison is primarily about how specifically we can modulate target emotion while minimizing unintended content changes, and Table~\ref{tab:main} indicates that contrastive-margin-based selector (CAS) best satisfies this requirement. 
We therefore adopt CAS as the default selector in subsequent experiments, as its margin-based assignment yields the most responsive and interpretable control behavior at inference time.

\subsubsection{Selection Rate}
\label{sec:top_r}

We next examine the selection rate $r$, i.e., the fraction of top-ranked neurons retained for  constructing ESN masks. By design, $r$ controls mask sparsity and balance between \emph{specificity} and \emph{coverage}: small $r$ retains only the most confident neurons, whereas large $r$ admits additional neurons that may encode weaker or non-specific signals \cite{tang-etal-2024-language}. Because ESN masks are subsequently used for inference-time intervention, robustness to $r$ is critical.

\begin{figure}[th!]
    \centering
    \begin{subfigure}[b]{0.33\linewidth}
        \centering
        \includegraphics[width=\linewidth]{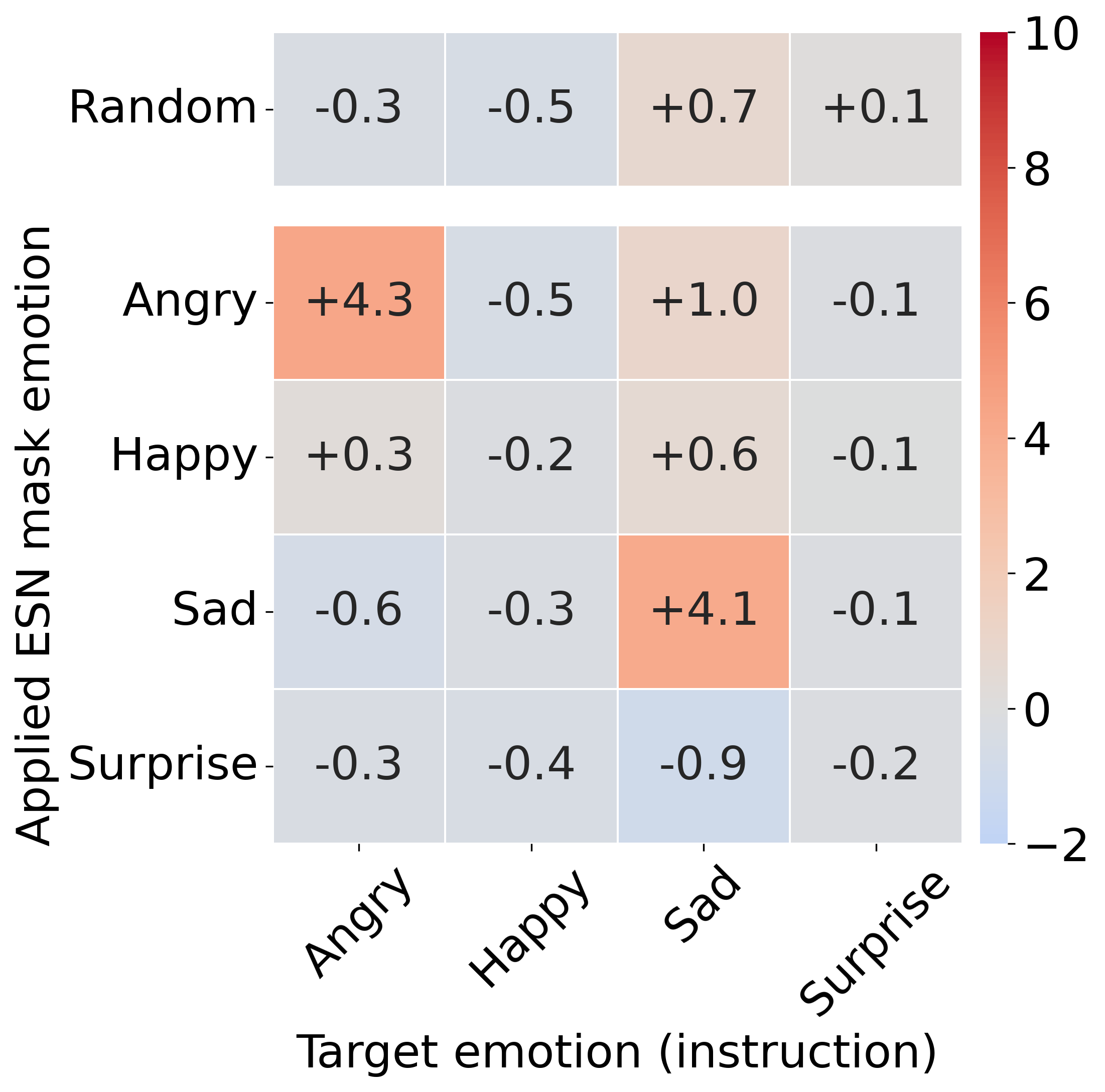}
        \caption{\textbf{$r=0.1\%$}}
    \end{subfigure}\hfill
    \begin{subfigure}[b]{0.33\linewidth}
        \centering
        \includegraphics[width=\linewidth]{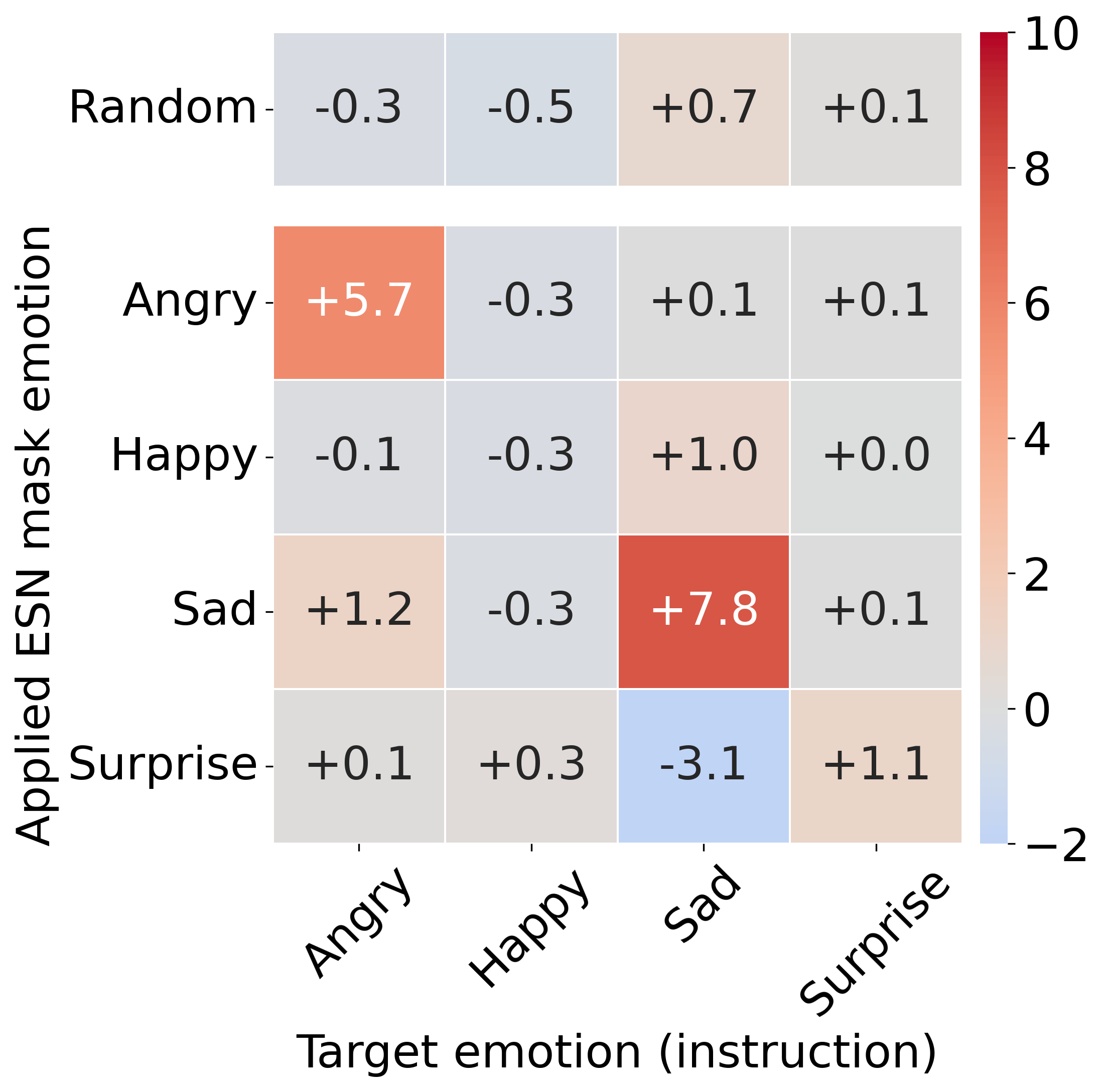}
        \caption{\textbf{$r=0.3\%$}}
    \end{subfigure}\hfill
    \begin{subfigure}[b]{0.33\linewidth}
        \centering
        \includegraphics[width=\linewidth]{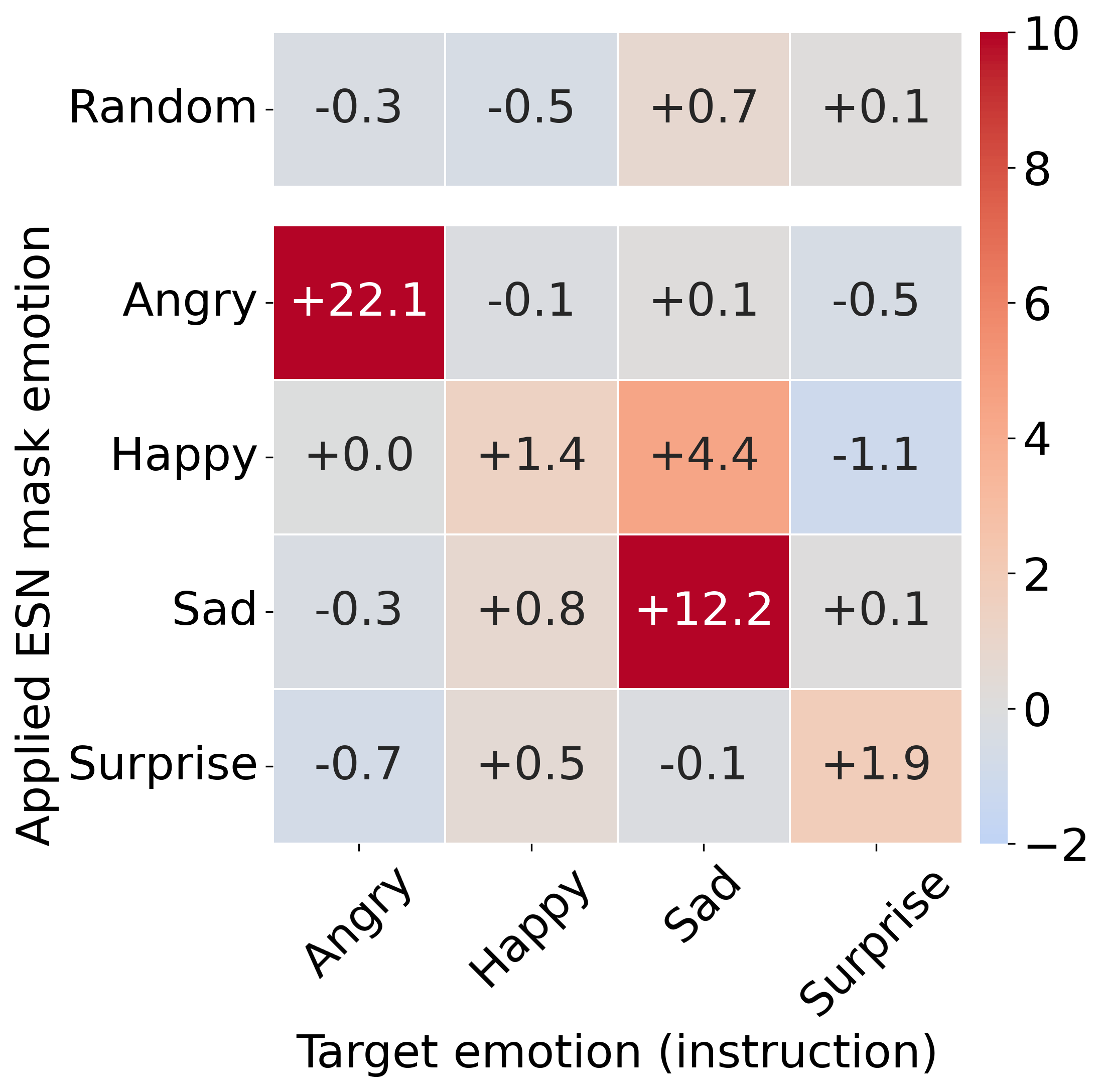}
        \caption{\textbf{$r=1\%$}}
    \end{subfigure}
    \vspace{-3mm}
    \caption{\textbf{Sensitivity to ESN selection rate $r$} (Qwen2.5-Omni-7B, CAS, $c{=}50$, steering, $\alpha{=}1.0$).}
    \label{fig:top_rate}
    \vspace{-3mm}
\end{figure}

Figure~\ref{fig:top_rate} suggests a largely monotonic trend within the tested range: with \(r{=}0.1\%\) the effect is present but weak and sparse (consistent with under-selection), while increasing \(r\) to a moderate range, \(0.3\%\) and \(0.5\%\) (in Figure~\ref{fig:steer_id_method}(d)) progressively strengthens the diagonal/self-effect and clarifies the specificity pattern. However, when \(r\) reaches \(1.0\%\), off-diagonal changes become more visible (e.g., +4.4 pp for Sad with a Happy-mask). This behavior supports that ESNs in LALMs are distributed rather than ultra-sparse, hence selecting too few neurons leaves part of the relevant signal unused. Since our sweep only extends to \(r{=}1.0\%\), we interpret this monotonicity as holding within the explored range and retain \(r{=}0.5\%\) as a practical default balancing compactness and effect strength.

\subsubsection{Success Set Size}
\label{success_size}
We then vary the success instance size $c$, the maximum number of filtered EVC successes retained per target emotion for activation aggregation. $c$ changes the amount of evidence used to estimate the neuron statistics in \S~\ref{sec:selectors}, so it is expected to affect the stability of ESN ranking.

\begin{figure}[ht!]
    \centering
    \begin{subfigure}[b]{0.33\linewidth}
        \centering
        \includegraphics[width=\linewidth]{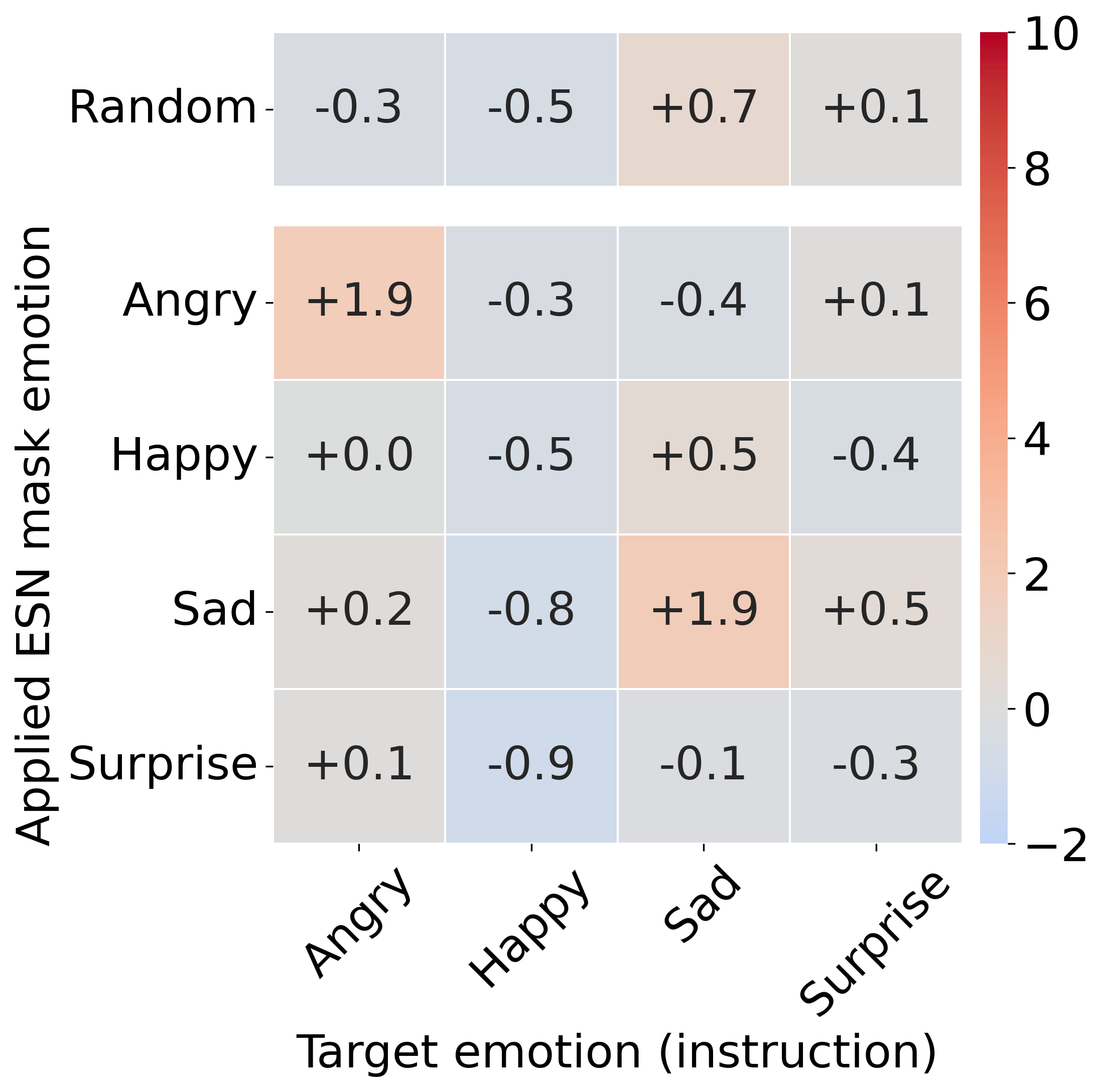}
        \caption{\textbf{$c{=}10$}}
    \end{subfigure}\hfill
    \begin{subfigure}[b]{0.33\linewidth}
        \centering
        \includegraphics[width=\linewidth]{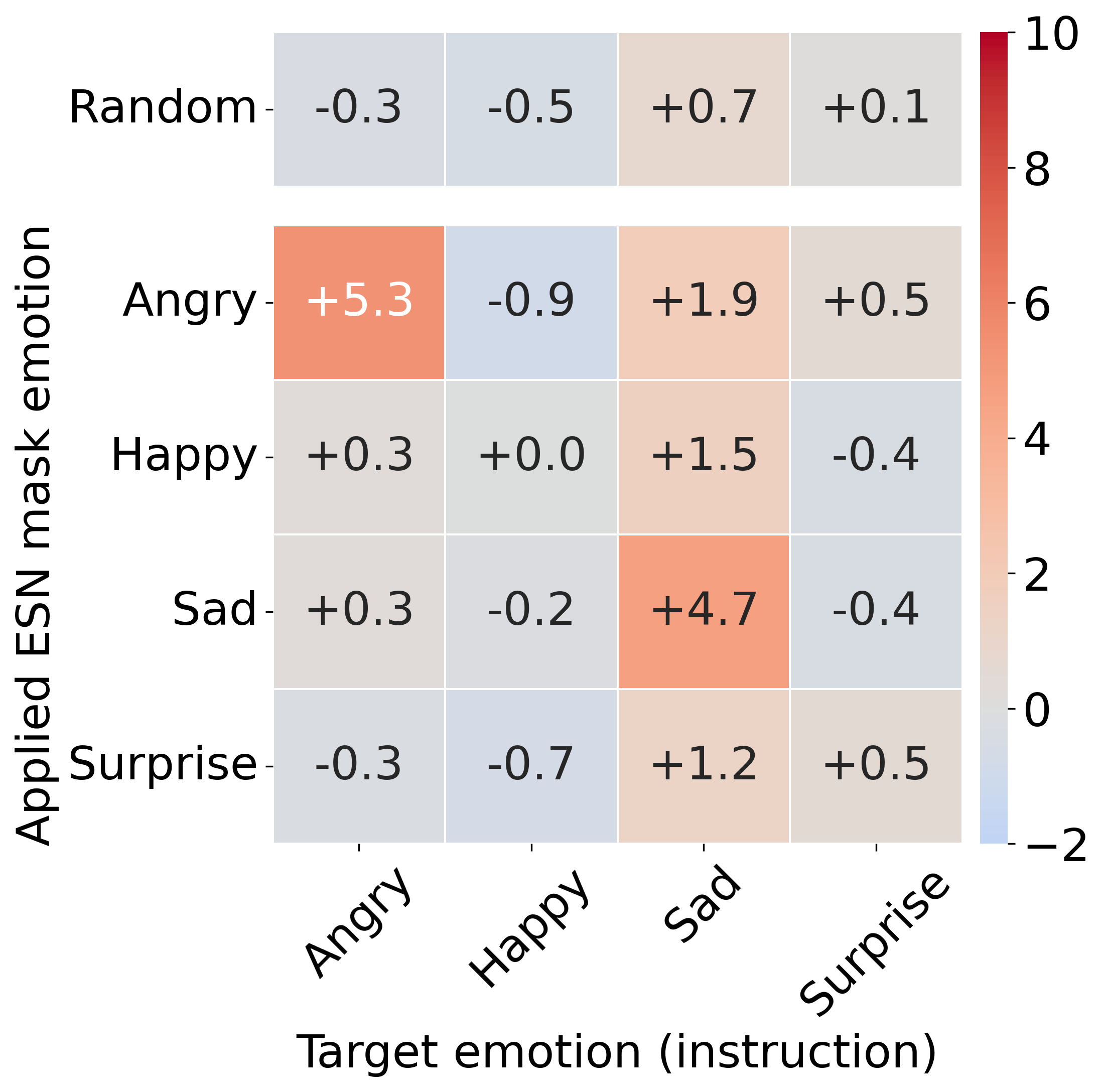}
        \caption{\textbf{$c{=}20$}}
    \end{subfigure}\hfill
    \begin{subfigure}[b]{0.33\linewidth}
        \centering
        \includegraphics[width=\linewidth]{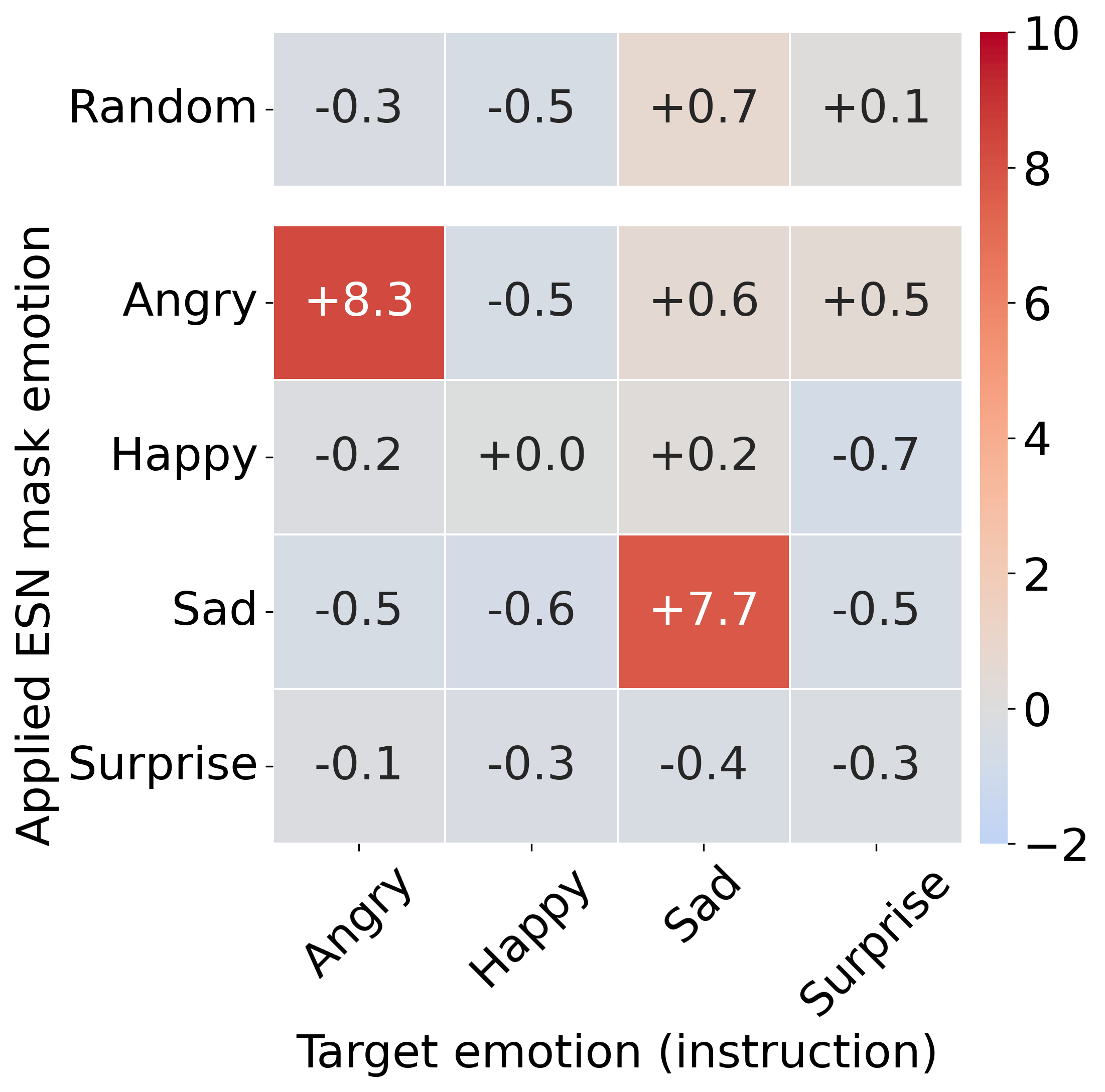}
        \caption{\textbf{$c{=}100$}}
    \end{subfigure}
    \vspace{-3mm}
    \caption{\textbf{Sensitivity to success instance size $c$} (Qwen2.5-Omni-7B, CAS, $r{=}0.5\%$, steering, $\alpha{=}1.0$). }
    \label{fig:success_instance}
    \vspace{-3mm}
\end{figure}

Figure~\ref{fig:success_instance} shows a clear sample-efficiency trend. With very small caps ($c{=}10$), the heatmaps are noisy and the diagonal is weak or inconsistent, indicating unstable ESN estimation from too few successful instances. As $c$ increases to $20$ or $50$ (Figure~\ref{fig:steer_id_method}(d)), the diagonal structure becomes noticeably clearer, suggesting that the selector now receives enough instances to retrieve more reliable emotion-conditioned activation patterns. While increasing further to $c{=}100$ does not improve self-effects proportionally, off-diagonal changes diminish, showing a more denoised emotion-specific neuron separation.

\subsection{Inference-Time Intervention Design}
Having established that identification parameters can affect ESNs' emotion control effects, we next analyze how inference-time intervention design affects the EVC outcome. We focus on two orthogonal choices: (1) intervention method and (2) intervention strength. 
\subsubsection{Intervention Methods}

\begin{figure}[ht!]
    \centering
    \begin{subfigure}[b]{0.33\columnwidth}
        \centering
        \includegraphics[width=\linewidth]{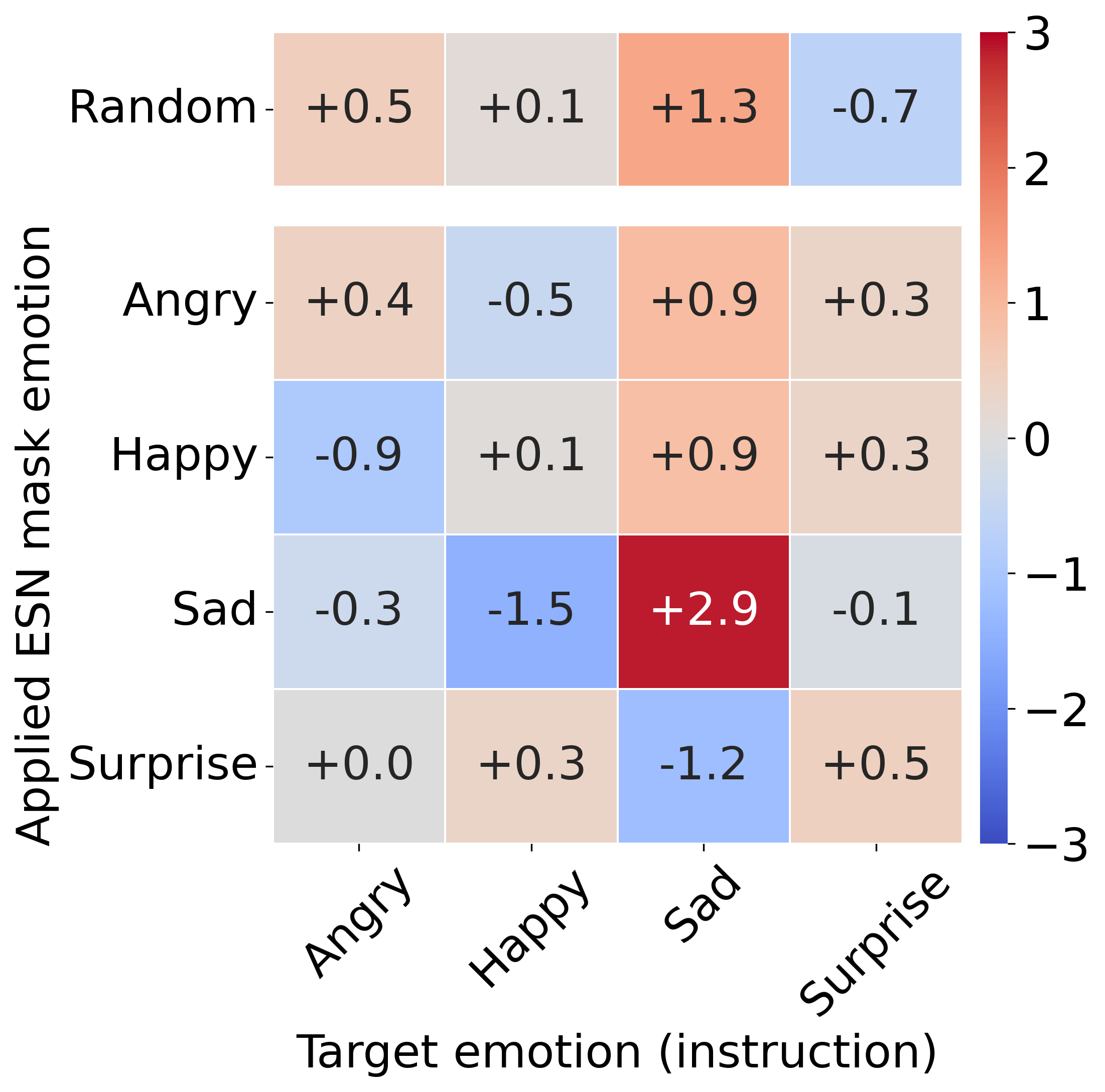}
        \caption{Add, \textbf{$\alpha$=0.30}}
    \end{subfigure}\hfill
    \begin{subfigure}[b]{0.33\columnwidth}
        \centering
        \includegraphics[width=\linewidth]{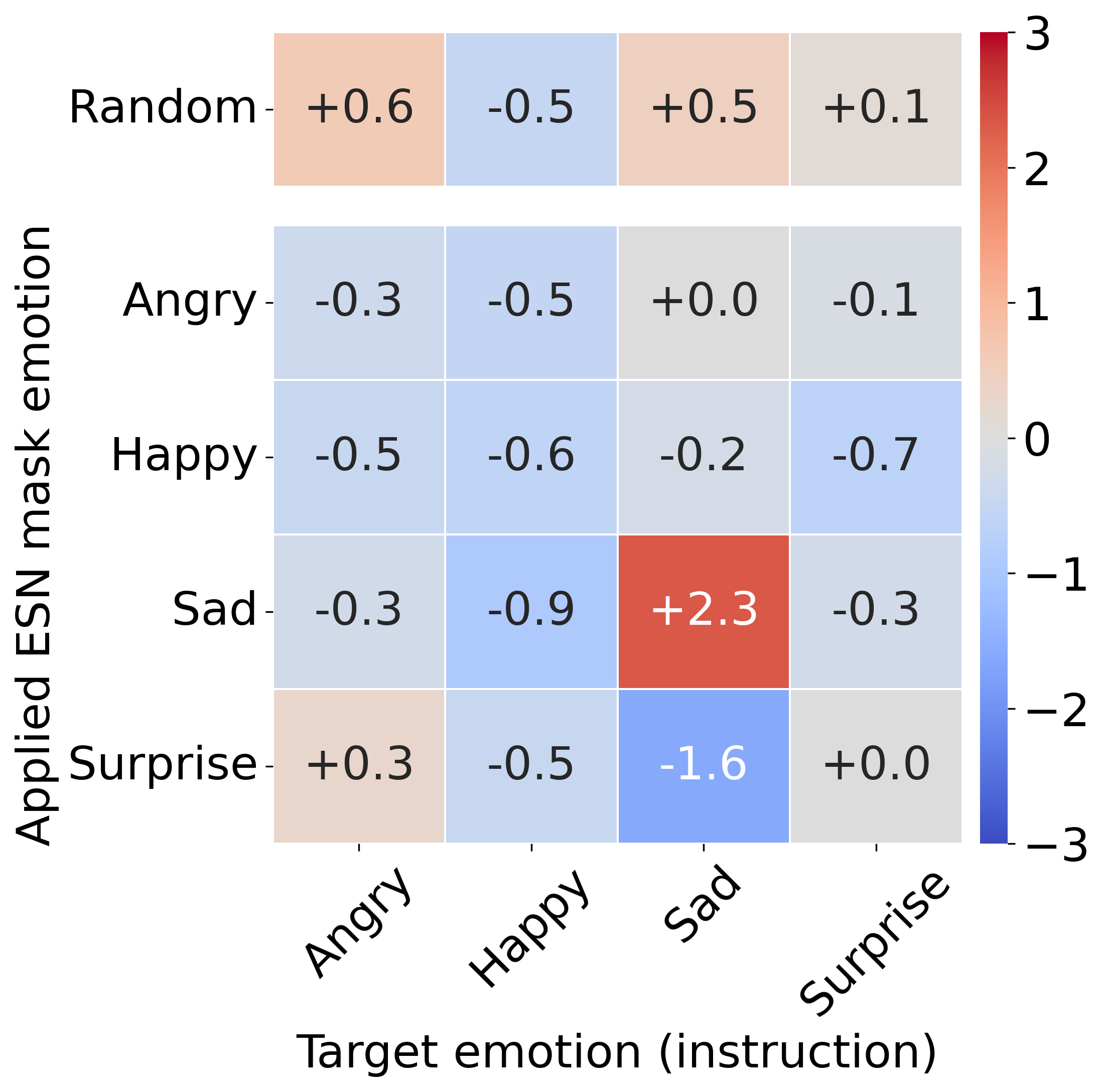}
        \caption{Clamp, \textbf{$\alpha$=0.10}}
    \end{subfigure}\hfill
    \begin{subfigure}[b]{0.33\columnwidth}
        \centering
        \includegraphics[width=\linewidth]{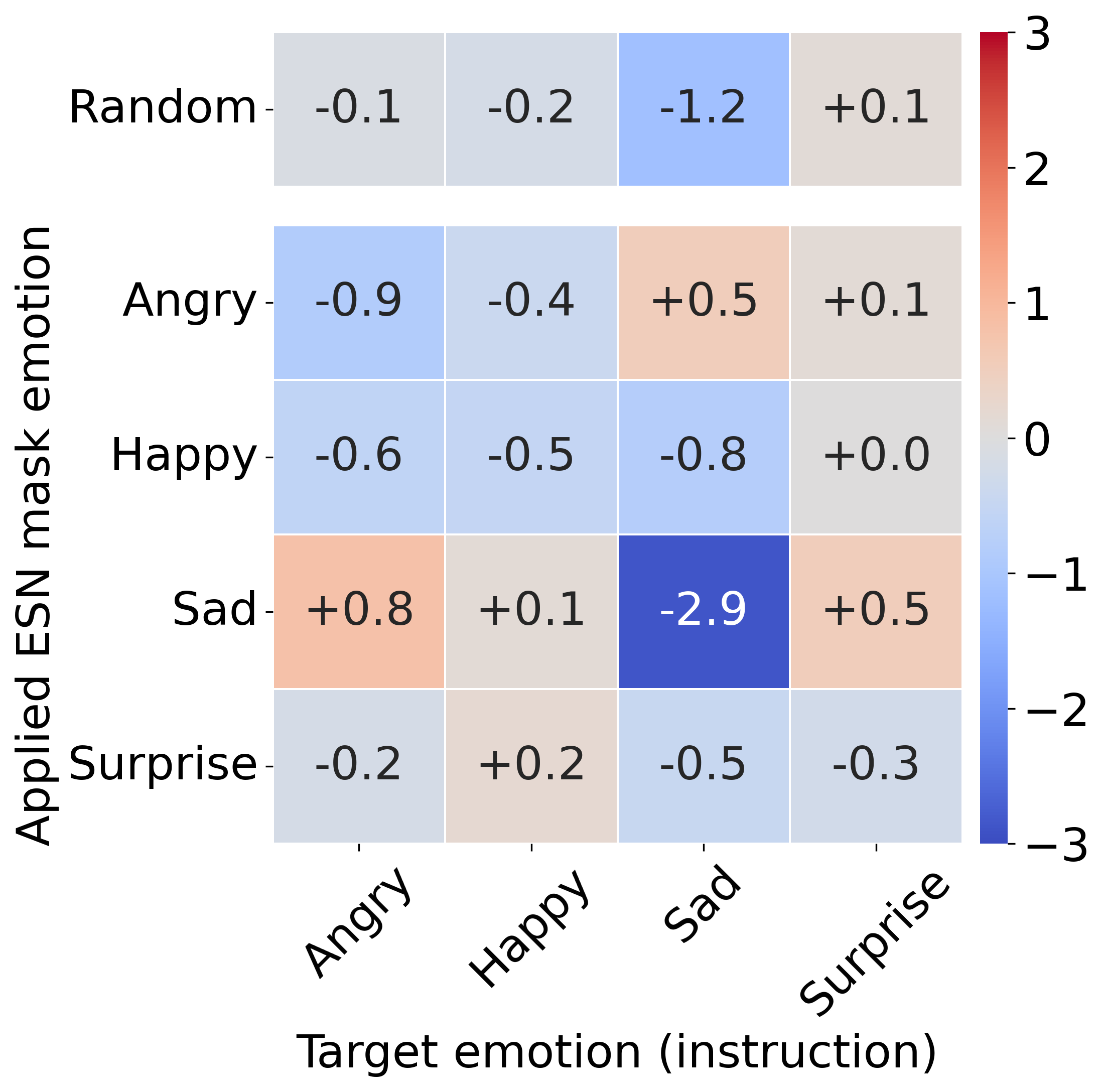}
        \caption{Deactivation}
    \end{subfigure}

    \vspace{-3mm}
    \caption{\textbf{Comparison of ESN intervention methods} (Qwen2.5-Omni-7B, CAS, $c{=}50$, $r{=}0.5\%$). As explained in \S~\ref{subsec:intervention}, $\alpha$ is method-specific, so we use different values to keep intervention strength roughly comparable across methods. Deactivation is parameter-free and hence reported without $\alpha$.} 
    \label{fig:interv_method}
    \vspace{-3mm}
\end{figure}

We further examine alternative ESN intervention methods beyond the scaling-style steering used in our main experiments. Figure~\ref{fig:interv_method} shows that, although additive and clamping interventions can induce partial target-emotion effects (e.g., a visible self-effect for sad under both methods), they do not yield consistent positive self-effects across emotions. 

The deactivation results provide two additional insights. First, for EVC, merely removing or suppressing non-target-emotion activations is often insufficient to reliably induce the desired expressive changes. Second, the consistently negative diagonal under deactivation offers additional evidence supporting the functional role of ESNs: when the identified neurons are silenced, target-emotion realization is impaired rather than improved. This pattern strengthens the interpretation that these neurons actively participate in emotion-related generation.
Based on this comparison, we adopt the best-performing gain-scaling steering intervention as the default in subsequent intervention strength analyses.

\subsubsection{Intervention Strength}
\label{sec:alpha}

\begin{table}[!ht]
\caption{\textbf{Emotion match and WER trade-off when increasing the intervention strength} (Qwen2.5-Omni-7B, CAS, $c{=}50$, $r{=}0.5\%$, steering).
}
\label{tab:alpha}
\centering
\resizebox{\columnwidth}{!}{%
\begin{tabular}{@{}lccccc@{}}
\toprule
         & \multicolumn{3}{c}{$\Delta$Emotion Match vs. Base (emotion2vec / Qwen3)}              &                             &                             \\ \cmidrule(lr){2-4}
$\alpha$ & Self-Effect $\uparrow$  & Cross-Effect Avg. $\downarrow$ & Self--Cross Gap $\uparrow$ & WER $\downarrow$ ($\Delta$) & UTMOS $\uparrow$ ($\Delta$) \\ \midrule
0.3      & +0.78 / +0.37           & +0.48 / -0.09                  & +0.30 / +0.45              & \textbf{19.33 (+0.33)}      & \textbf{4.00 (0.00)}        \\
0.5      & +2.00 / +1.13           & -0.11 / \textbf{-0.22}         & +2.11 / +1.36              & 20.67 (+1.67)               & \textbf{4.00 (0.00)}        \\
1.0      & +4.27 / +3.60           & +0.34 / -0.05                  & +3.93 / +3.65              & 29.65 (+10.65)              & 3.99 (-0.01)                \\
2.0      & \textbf{+8.89 / +20.67} & \textbf{-0.14} / +0.02         & \textbf{+9.02 / +20.64}    & 203.65 (+184.65)            & 3.91 (-0.09)                \\ \bottomrule
\end{tabular}%
}
\vspace{-3mm}
\end{table}

\begin{figure}[th]
    \centering
    \begin{subfigure}[b]{0.33\linewidth}
        \centering
        \includegraphics[width=\linewidth]{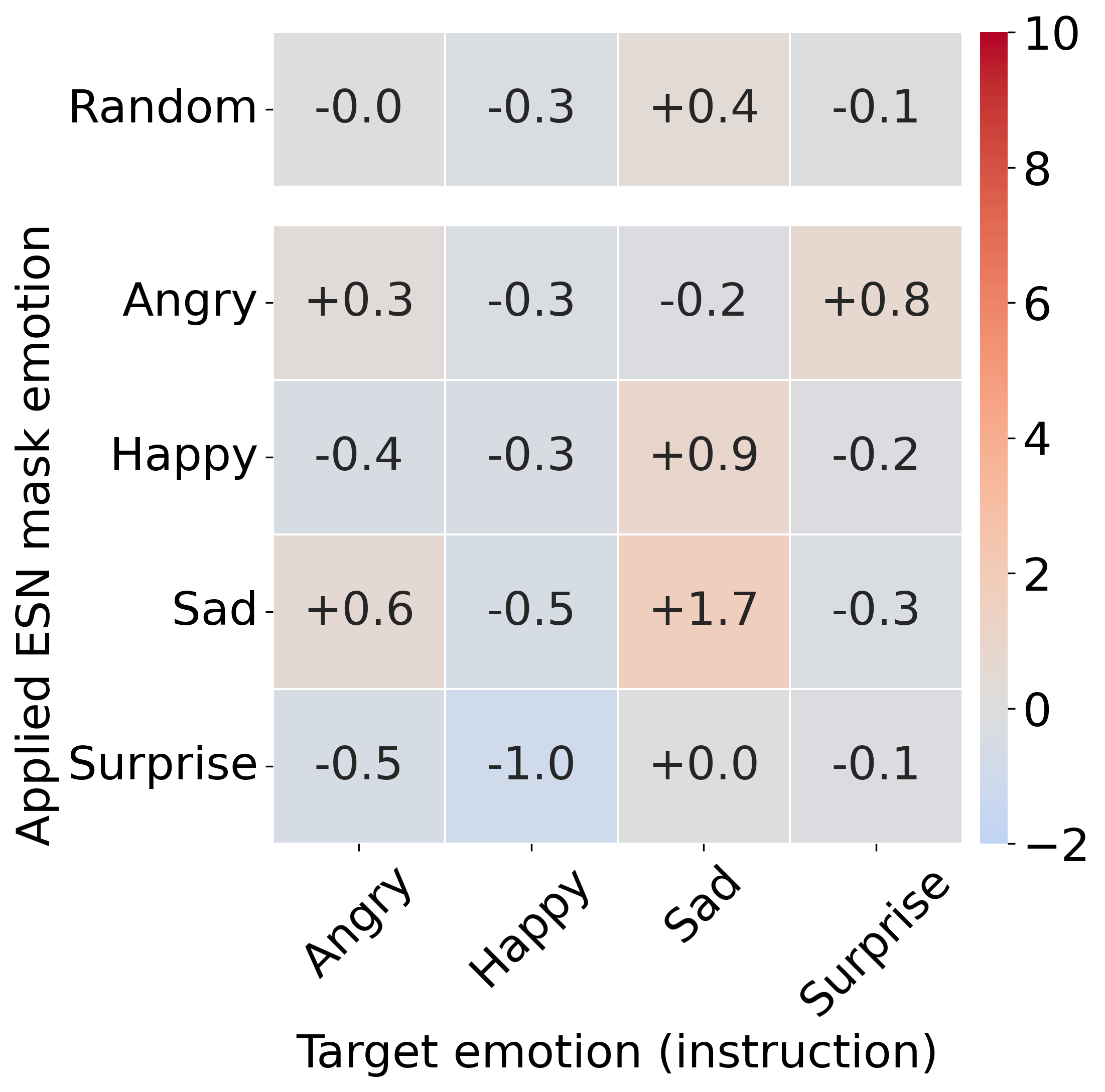}
        \caption{\textbf{$\alpha$=0.30}}
    \end{subfigure}\hfill
    \begin{subfigure}[b]{0.33\linewidth}
        \centering
        \includegraphics[width=\linewidth]{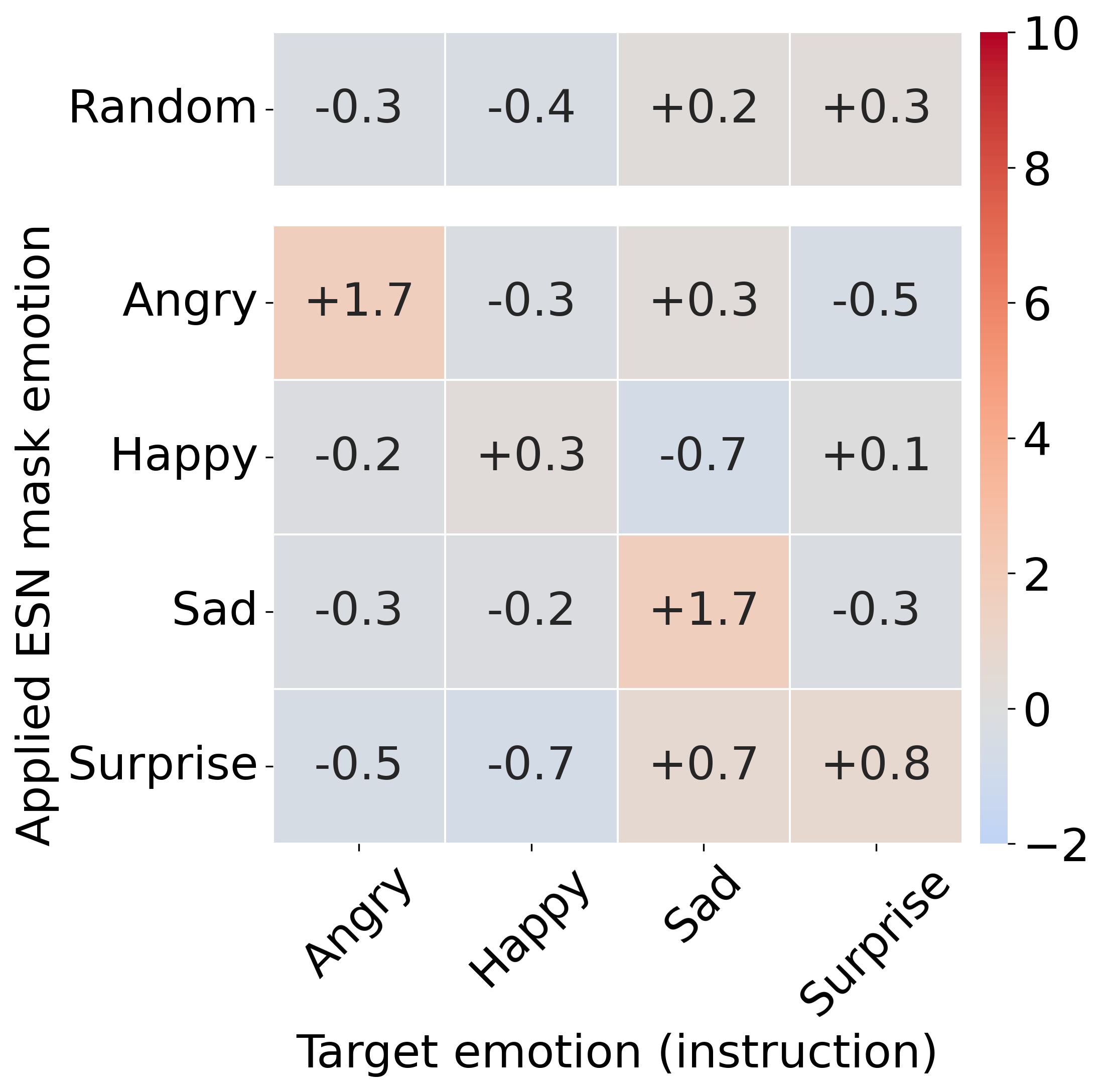}
        \caption{\textbf{$\alpha$=0.50}}
    \end{subfigure}\hfill
    \begin{subfigure}[b]{0.33\linewidth}
        \centering
        \includegraphics[width=\linewidth]{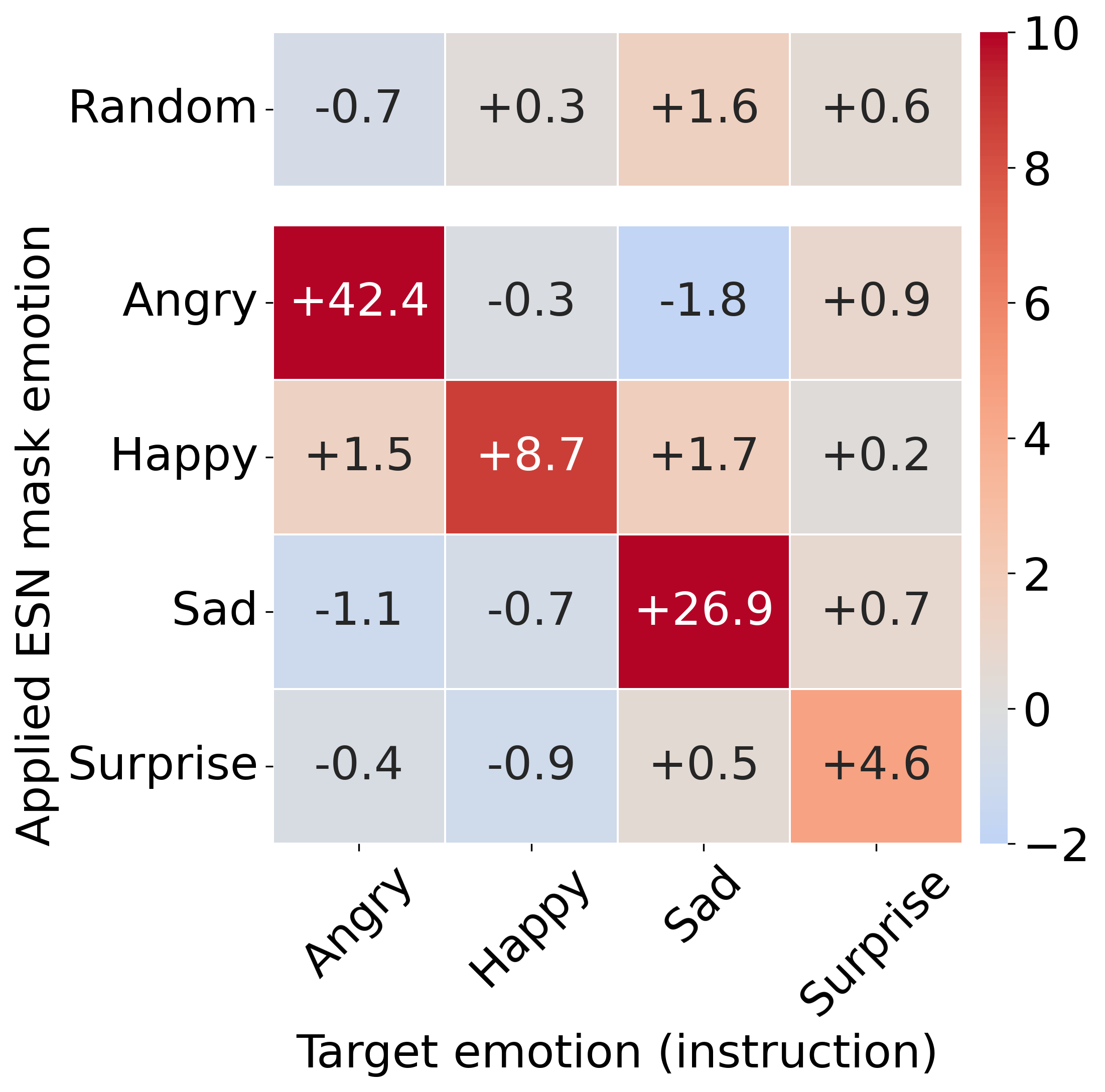}
        \caption{\textbf{$\alpha$=2.00}}
    \end{subfigure}
    \vspace{-3mm}
    \caption{\textbf{Sensitivity to intervention strength $\alpha$} (Qwen2.5-Omni-7B, CAS, $c{=}50$, $r{=}0.5\%$, steering).}
    \label{fig:steer_alpha}
    \vspace{-4mm}
\end{figure}

Taking activation steering as a probe method, we report the sensitivity of intervention effects to the intervention strength $\alpha$ in 
Table~\ref{tab:alpha} and Figure~\ref{fig:steer_alpha}. Increasing $\alpha$ consistently strengthens the intended self-effect on emotion match, as reflected in progressively darker diagonal cells in the heatmaps. Off-diagonal cross-effect, on the other hand, does not monotonically expand with $\alpha$ but rather fluctuates, leading to widening self--cross gaps.

The largest $\alpha$ further highlights the trade-off with content preservation. While $\alpha{=}2.0$ yields the strongest emotion match gains (+8.89/+20.67 pp) and a large self--cross gap (+9.02/+20.64 pp), it catastrophically harms content preservation (WER 203.65\%, +184.65 pp over baseline) and slightly lowers naturalness. In contrast, in this setting, moderate steering ($\alpha{=}0.3$--$0.5$) keeps WER near the baseline and leaves UTMOS unchanged, while still providing measurable improvements in target emotion match.

\subsection{Human Evaluation and Human-Supervised ESN Identification}
\label{sec:human}

\begin{figure}[ht!]
    \centering
        \includegraphics[width=0.8\columnwidth]{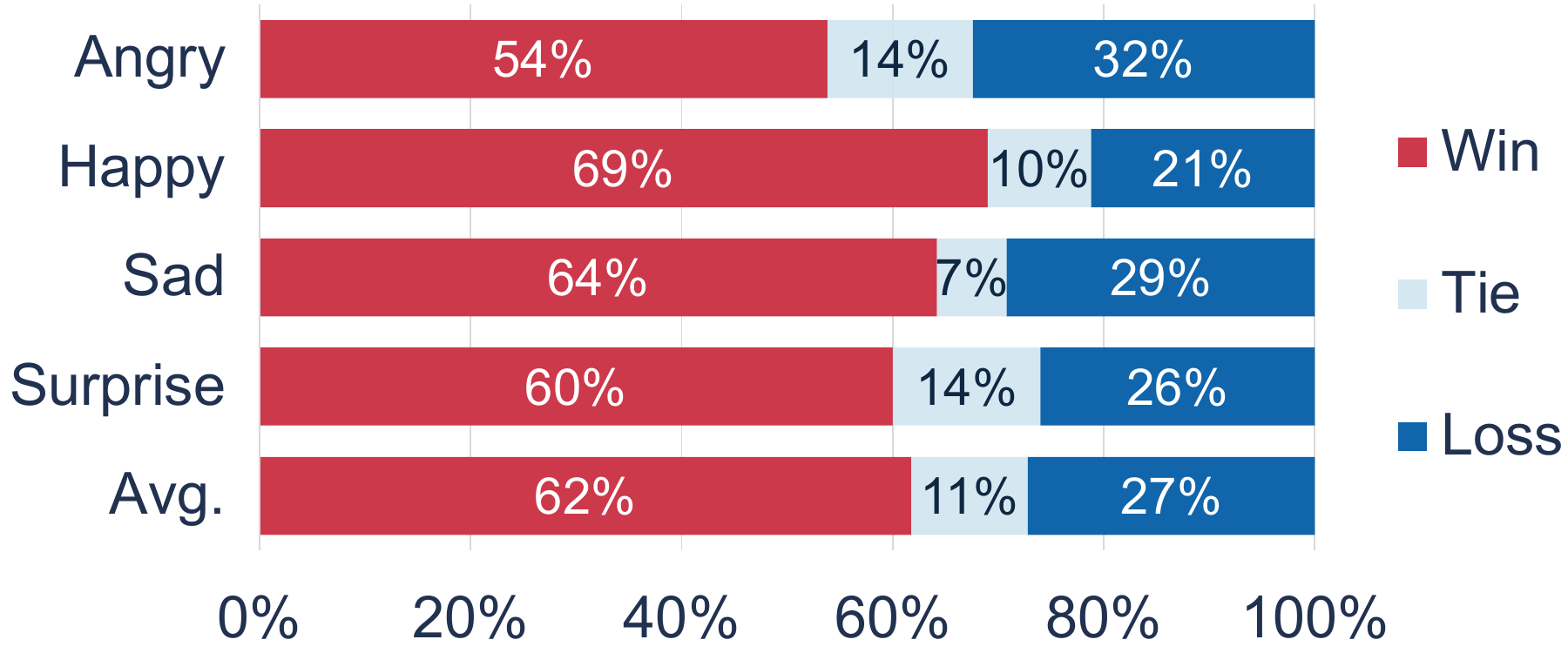}
        \caption{\textbf{Human listening evaluation under the anchor intervention setting.} Win/Tie/Loss rates comparing the intervened system (Qwen2.5-Omni-7B, CAS, $c{=}50$, $r{=}0.5\%$, steering, $\alpha{=}1.0$) against the unintervened baseline. 20 participants.}
    \label{fig:human_eval_winrate}
    \vspace{-3mm}
\end{figure}

To complement automatic SER evaluation, we conduct a human listening study under our anchor configuration, matching the SER-filtered setting in Figure~\ref{fig:steer_id_method}(d) for direct comparability.
In each trial, participants listen to a pair of speech samples and select the clip that better matches a specified target emotion; an ``indecisive'' option is available if the difference is unclear. Each pair consists of (1) the baseline output and (2) the corresponding output generated with identical source speech and instruction under an ESN intervention. Each participant was provided with the same test set of 100 sample pairs (25 per emotion). Samples are randomized and volume-normalized, and raters are blind to model identity. We collected valid responses from 20 participants proficient in English. Listeners were aged 21--32 years; 7 identified as female, 13 as male. All reported normal hearing. Results are summarized in Figure~\ref{fig:human_eval_winrate}: listeners prefer the intervened system in all four emotions, with average win/tie/loss rates of 62\%/11\%/27\%. Gains are most pronounced for happy (69\% win) and remain strong for sad and surprise, while angry is still favorable (54\% win), supporting that ESN steering yields perceptible emotion shifts that are also reflected in human listening judgments.

\begin{table}[ht]
\caption{\textbf{Similarity between ESNs identified from human- vs.\ SER-filtered instances}, measured in Jaccard overlap, direct match rate, and Jensen--Shannon divergence (JSD).}
\label{tab:human_id_overlap}
\centering
\resizebox{0.6\columnwidth}{!}{%
    \begin{tabular}{lccc}
    \toprule
    Emotion  & Jaccard & Direct Match & JSD \\ \midrule
    Angry    & 0.217   & 35.71\% & 0.009 \\
    Happy    & 0.075   & 13.95\% & 0.040 \\
    Sad      & 0.109   & 19.57\% & 0.010 \\
    Surprise & 0.071   & 13.27\% & 0.025 \\ \midrule
    Average  & 0.118   & 20.63\% & 0.021 \\ \bottomrule
    \end{tabular}%
    }
\vspace{-3mm}
\end{table}

To assess the consistency between automatic and human supervision during ESN identification, we further compare neuron sets obtained from SER-filtered versus human-filtered success instances. Table~\ref{tab:human_id_overlap} shows that overlap at the neuron level is partial, indicating that the two filtering strategies select different specific units. However, the layer-wise distributions between the two are highly similar (average JSD 0.021), suggesting that both supervision signals concentrate ESNs in similar MLP layers. Agreement is strongest for angry (Jaccard 0.217, JSD 0.009), whereas happy and surprise exhibit lower overlap, consistent with their weaker or more ambiguous emotion signals.

The CAS-H row in Table~\ref{tab:main} replaces SER-filtered successes with human-judged ones for ESN identification, and yields the same qualitative behavior: positive self-effects with limited cross-emotion spillover. Compared to CAS, CAS-H slightly reduces self-effect and self--cross gap, but also substantially lowers $\Delta$WER, suggesting that human supervision preferentially retains content-faithful successes and produces a more conservative yet robust ESN set for controllable EVC.

\subsection{Localization of Emotion-Sensitive Neurons}
\label{sec:location}

\begin{figure}[ht!]
    \centering
    \begin{subfigure}[b]{0.32\columnwidth}
        \centering
        \includegraphics[width=\linewidth]{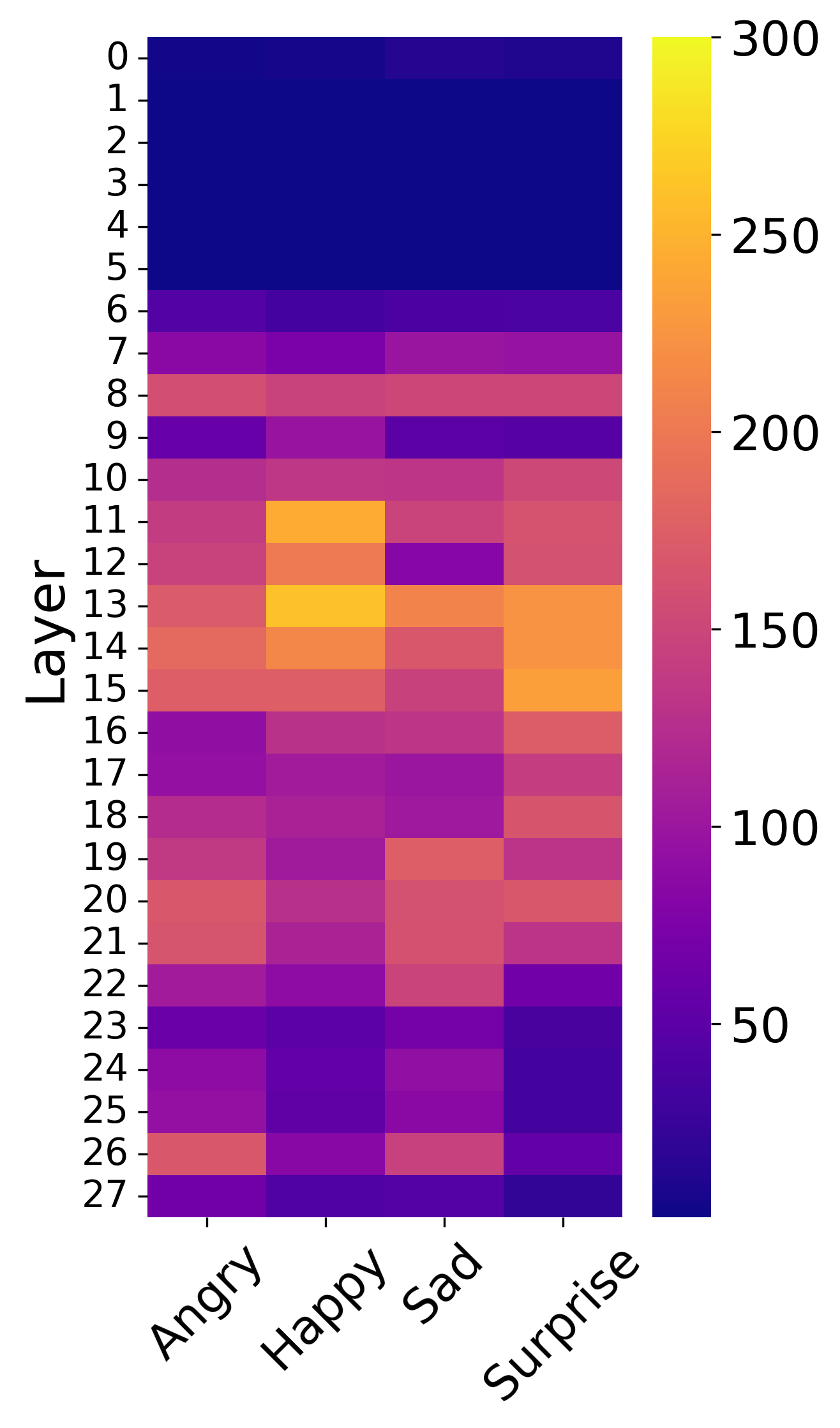}
        \caption{Qwen2.5-Omni-7B}
    \end{subfigure}\hfill
    \begin{subfigure}[b]{0.32\columnwidth}
        \centering
        \includegraphics[width=\linewidth]{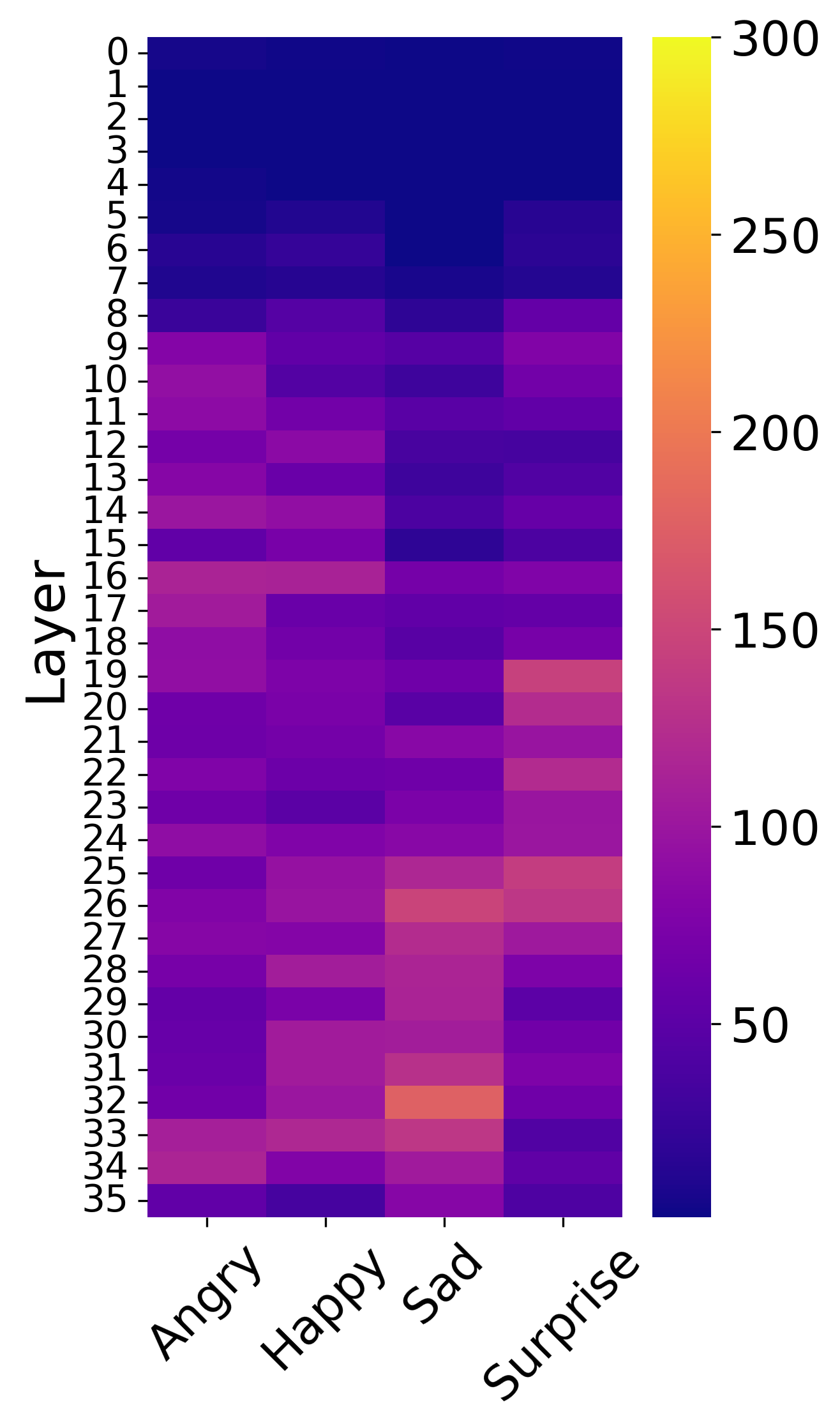}
        \caption{MiniCPM-o 4.5}
    \end{subfigure}\hfill
    \begin{subfigure}[b]{0.32\columnwidth}
        \centering
        \includegraphics[width=\linewidth]{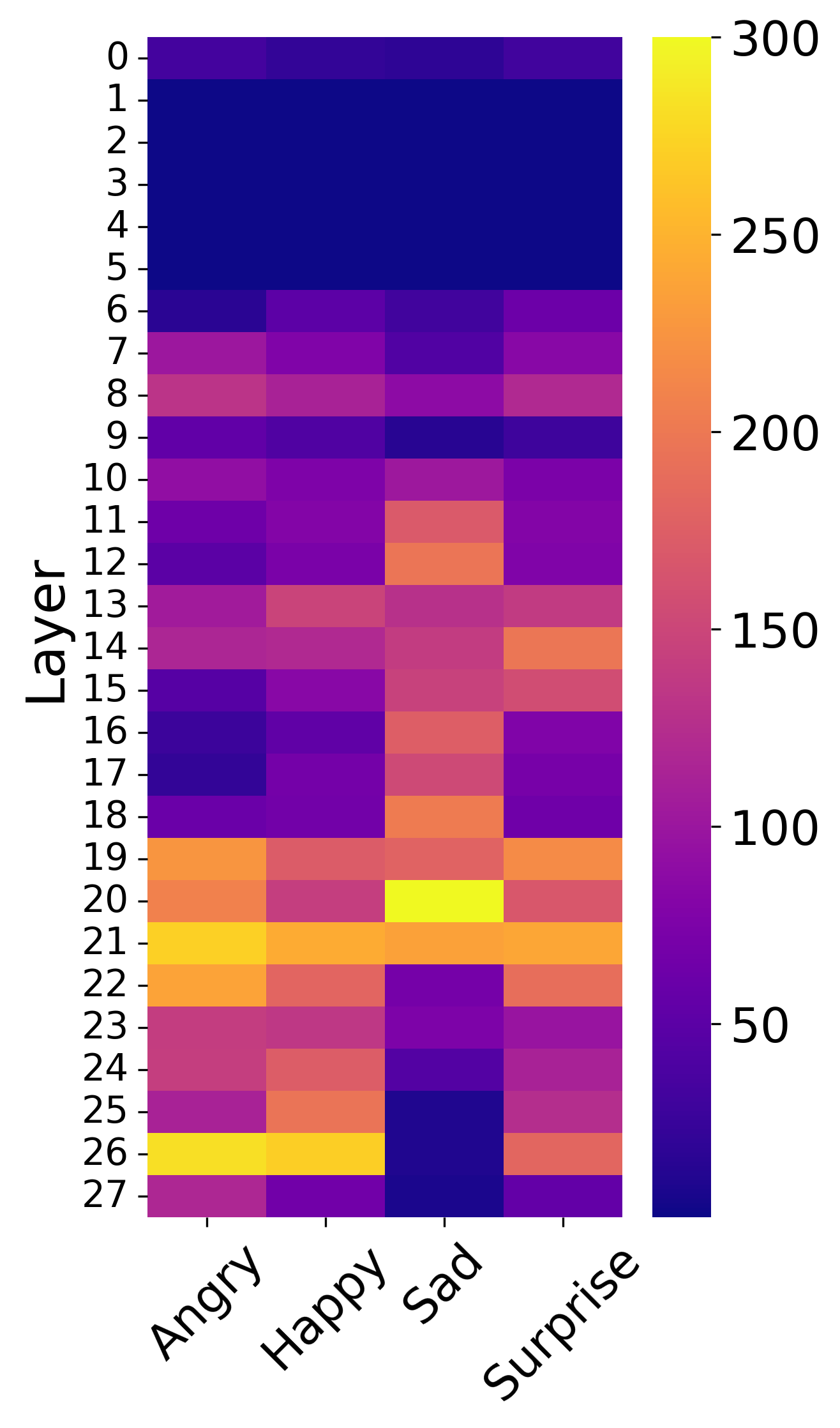}
        \caption{Kimi-Audio}
    \end{subfigure}
    \vspace{-2mm}
    \caption{\textbf{Layer-wise distribution of CAS-identified ESNs across three LALMs} ($r{=}0.5\%$). Heatmaps show the number of selected ESNs per decoder layer and target emotion. Qwen2.5-Omni-7B and Kimi-Audio use 28-layer decoders, while MiniCPM-o 4.5 uses a 36-layer decoder. Values indicate where selected neurons are concentrated, not intervention effect size.}
    \vspace{-3mm}
    \label{fig:layers}
\end{figure}

We then visualize the concentrations of the ESNs across the MLP layers in Figure~\ref{fig:layers}, which shows that ESNs are not uniformly distributed. Across all three LALMs, they concentrate primarily in intermediate-to-late layers of the language-model decoder, with relatively few units selected in the earliest layers. For Qwen2.5-Omni-7B, ESNs cluster most strongly in approximately layers 11--15, with pronounced peaks for happy and surprise. MiniCPM-o 4.5 exhibits a more diffuse distribution but still places greater mass in the latter half of the decoder. Kimi-Audio shows the strongest late-layer concentration, particularly in layers 19--22, with marked peaks for happy and angry in layer 26. This consistent pattern suggests that, in the tested models, emotion-relevant control signals are more prominently captured in intermediate-to-late language-model layers than in earlier layers.

\begin{table}[!ht]
\caption{\textbf{Activation steering in the speech synthesis module of Qwen2.5-Omni-7B.}
Results obtained by applying the same settings ($c{=}50$, $r{=}0.5\%$, $\alpha{=}1.0$) as in Table~\ref{tab:main}, but identified and intervening on the synthesis module MLP layers.
}
\label{tab:tts}
\centering
\resizebox{\columnwidth}{!}{%
\begin{tabular}{@{}lccccc@{}}
\toprule
         & \multicolumn{3}{c}{$\Delta$Emotion Match vs. Base (emotion2vec / Qwen3)}             &                             &                             \\ \cmidrule(lr){2-4}
Selector & Self-Effect $\uparrow$ & Cross-Effect Avg. $\downarrow$ & Self--Cross Gap $\uparrow$ & WER $\downarrow$ ($\Delta$) & UTMOS $\uparrow$ ($\Delta$) \\ \midrule
Random   & +0.52 / +0.02          & --                             & --                         & 19.21 (+0.21)               & 4.01 (+0.01)                \\
LAP      & \textbf{+0.32} / -0.08 & +0.12 / +0.23                  & \textbf{+0.20} / -0.32     & \textbf{19.00 (0.00)}       & 3.96 (-0.04)                \\
LAPE     & -0.87 / -0.22          & \textbf{-0.13 / -0.27}         & -0.73 / \textbf{+0.06}     & \textbf{19.00 (0.00)}       & \textbf{4.01 (+0.01)}       \\
MAD      & -0.68 / \textbf{0.08}  & +0.51 / +0.13                  & -1.19 / -0.04              & \textbf{19.00 (0.00)}       & 4.00 (0.00)                 \\
CAS      & +0.12 / -0.42          & +0.45 / -0.19                  & -0.33 / -0.22              & 19.19 (+0.19)               & 3.99 (-0.01)                \\ \bottomrule
\end{tabular}%
}
\vspace{-3mm}
\end{table}

We also explore an alternative module that could potentially contain ESNs. As demonstrated in Table~\ref{tab:tts}, repeating the same identification-intervention pipeline in the speech synthesis MLP of Qwen2.5-Omni-7B under identical settings eliminates the clean positive self-effects observed in the language-model MLPs. Across selectors, self--cross gaps become near-zero or negative under both SER judges, regardless of selector. Minimal changes in WER and UTMOS further indicate that synthesis side interventions are largely inert rather than emotion-specific.
These results suggest that language-model decoder MLPs yield the most actionable ESNs, rather than the downstream synthesis modules. Practically, this means that monitoring and intervening at the language-model side feed-forward blocks in LALMs is a more effective strategy for controllable affect steering than targeting the synthesis MLPs directly.

\subsection{Do Intervention Effects Generalize to Unseen Speakers?}

\begin{figure}[ht!]
    \centering
        \includegraphics[width=0.9\linewidth]{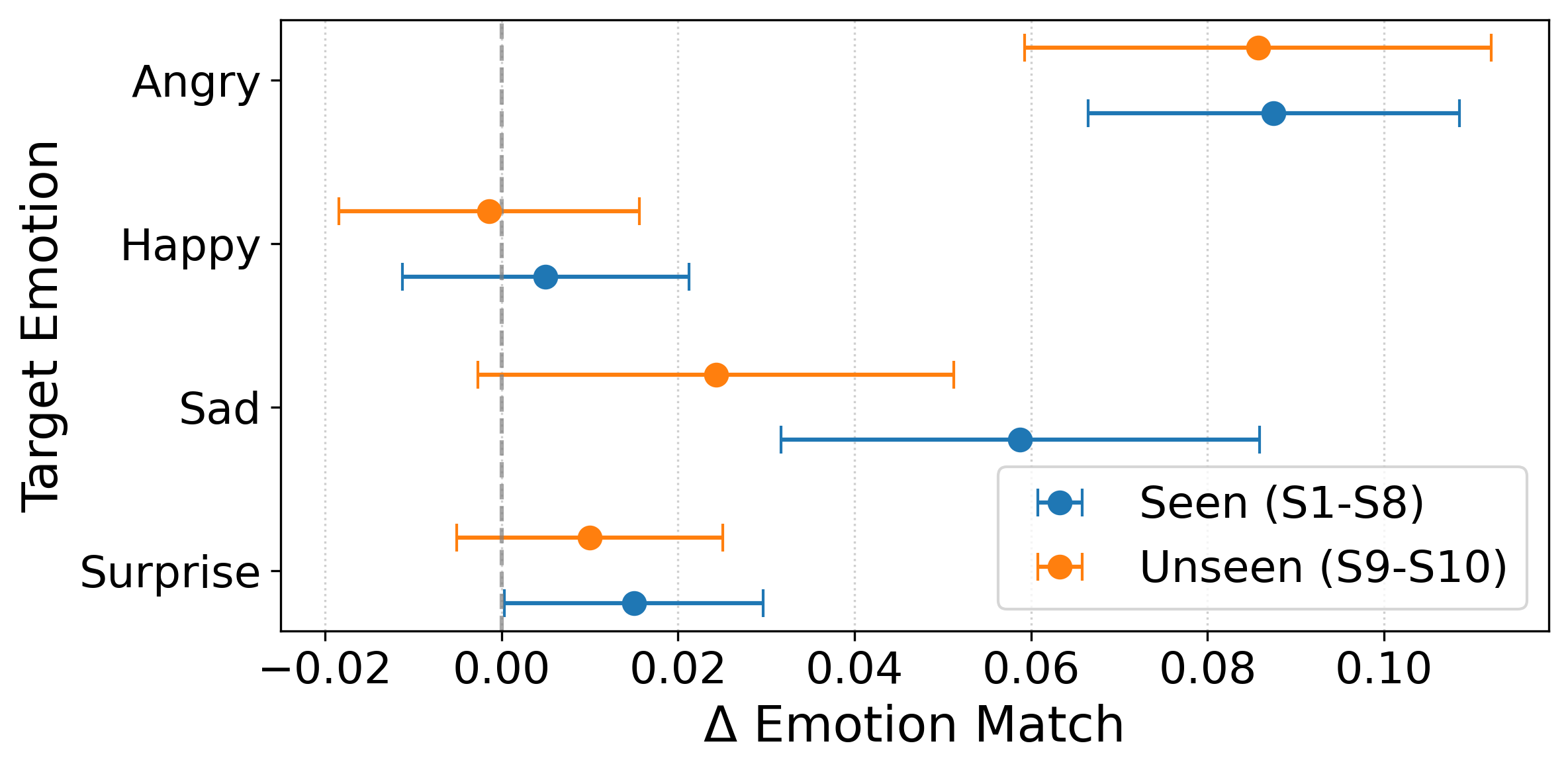}
        \vspace{-3mm}
        \caption{\textbf{Intervention effect on seen and unseen speakers.} $\Delta$ Emotion match size changes (pp) of intervention (Qwen2.5-Omni-7B, CAS, $c{=}50$, $r{=}0.5\%$, steering, $\alpha{=}1.0$) in comparison with the unintervened baseline are shown for seen speakers (\textsc{Test-seen} split) and unseen speakers (\textsc{Test-unseen} split), with 95\% confidence intervals.}
    \label{fig:seen_unseen}
    \vspace{-3mm}
\end{figure}

Finally, we evaluate whether ESN-based control generalizes beyond speakers used during neuron identification. Figure~\ref{fig:seen_unseen} shows that the intervention effect transfers to unseen speakers, but the magnitude of transfer is emotion-dependent. For angry, gains remain strong in both seen and unseen conditions, indicating a relatively speaker-robust emotion signal. Surprise also maintains positive effects across splits, albeit at reduced magnitude. In contrast, sad exhibits a noticeable drop from seen to unseen speakers, suggesting partial speaker-specificity in the learned signal. Happy shows the weakest transfer, with modest gains on seen speakers and near-zero or slightly negative effects on unseen speakers. 
A practical implication is that speaker-robust ESN control is feasible, but likely benefits from more diverse identification data.

\section{Conclusion}
We presented a systematic neuron-level investigation of emotion control in speech-generative LALMs under an emotional voice conversion setting.
Our results suggest that compact emotion-sensitive neuron sets can be identified, are interventionally actionable, and can be leveraged to steer emotional expressiveness at inference time without any weight updates. Crucially, we showed that reliable identification of ESNs requires success-filtered activation aggregation that respects the multi-objective nature of EVC. Contrastive-margin- and mean-deviation-based selectors consistently outperform frequency- and entropy-based heuristics, yielding stronger self–cross separation and more emotion-specific control. 
We further demonstrated a key trade-off between emotional expressiveness and content preservation: moderate interventions amplify target-emotion evidence, while excessive steering increases WER through semantic drift rather than a loss of intelligibility, as naturalness remains largely stable.
Localization analyses reveal that actionable emotion-control signals concentrate in intermediate-to-late decoder MLP layers on the language-model side, while analogous interventions in downstream speech synthesis modules are largely inert. This finding clarifies where emotion control is actually mediated in speech-generative LALMs and narrows the locus for future interpretability efforts.

Together, these results establish a practical and mechanistically grounded pipeline for \emph{training-free emotion control} in speech generation. Beyond emotional voice conversion, our framework opens the door to fine-grained, inference-time manipulation of paralinguistic attributes in multimodal generative systems without re-training or task-specific fine-tuning.

\section{Acknowledgments}
We thank the National Science Foundation (NSF) for support under CAREER Award IIS-2533652.

\section{Generative AI Use Disclosure}
Generative AI tools were employed solely for language polishing of text written by the authors. These tools were not used to generate scientific content, results, experimental designs, analyses, or conclusions. All authors are responsible for the full content of this paper and consent to its submission.

\bibliographystyle{IEEEtran}
\bibliography{mybib}

\end{document}